\pdfoutput=1

\documentclass[11pt]{article}

\usepackage{acl}

\usepackage{times}
\usepackage{latexsym}

\usepackage[T1]{fontenc}

\usepackage[utf8]{inputenc}

\usepackage{microtype}

\usepackage{tikz}
\pgfdeclarelayer{background}
\pgfdeclarelayer{main}
\pgfsetlayers{background,main}

\usetikzlibrary{calc}
\usepackage{graphicx}
\usepackage{color}
\usepackage{url}
\usepackage{rotating}

\definecolor{dgreen}{rgb}{0.0,0.75,0.0}

\newcommand*{\minWidth}{35}
\newcommand*{\maxValue}{1}
\newcommand{\makeBar}[3]{%
    \hspace{-6.5pt}
  \tikz[baseline]{
    \node[anchor=base,text width=\minWidth,align=#2,inner sep=0pt,inner xsep=\tabcolsep,outer sep=0pt] (n) {\strut{#1}};
    \begin{pgfonlayer}{background}
        {
       \edef\color{#3}
       \pgfmathparse{abs(#1/\maxValue)}
       \edef\contents{{\pgfmathresult}}
       \fill[font=\boldmath,color=\color] (n.north west) rectangle ($(n.south west)!{{\contents}}!(n.south east)$);
       }
    \end{pgfonlayer}
  }
    \hspace{-9.5pt}
}

\newcommand{\makeDataBar}[1]{\makeBar{#1}{center}{cyan!25}}
\newcommand{\makeAverageBar}[1]{\makeBar{#1}{center}{green!25}}

\title{Making Transformers Solve Compositional Tasks}

\author{Santiago Onta\~{n}\'{o}n, Joshua Ainslie, Vaclav Cvicek, Zachary Fisher \\
  Google Research \\
  \{\texttt{santiontanon, jainslie, vcvicek, zachfisher}\}\texttt{@google.com}}

\begin{document}
\maketitle
\begin{abstract}
Several studies have reported the inability of Transformer models to generalize compositionally, a key type of generalization in many NLP tasks such as semantic parsing. In this paper we explore the design space of Transformer models showing that the inductive biases given to the model by several design decisions significantly impact compositional generalization. We identified Transformer configurations that generalize compositionally significantly better than previously reported in the literature in many compositional tasks. We achieve state-of-the-art results in a semantic parsing compositional generalization benchmark (COGS), and a string edit operation composition benchmark (PCFG).
\end{abstract}

\section{Introduction}\label{sec:intro}

Although modern neural network architectures reach state-of-the-art performance in many challenging natural language tasks, they seem to exhibit a low amount of ``compositional generalization'', i.e., the ability to learn a set of basic primitives and combine them in more complex ways than those seen during training~\cite{hupkes2020compositionality}. For example, suppose a system has learned the meaning of ``jump'' and that ``jump twice'' means that the action ``jump'' has to be repeated two times. Upon learning the meaning of the action ``jax'', it should be able to infer what ``jax twice'' means. Compositional generalization is a key aspect of natural language and many other tasks we might want machine learning models to learn.

While both humans and classical AI techniques (such as grammars or search-based systems) can handle compositional tasks with relative ease, it seems that modern deep learning techniques do not possess this ability. A key question is thus: Can we build deep learning architectures that can also solve compositional tasks? In this paper we focus on Transformers~\cite{vaswani2017attention}, which have been shown in the literature to exhibit poor compositional generalization (see Section \ref{sec:background}). Through an empirical study, we show that this can be improved. 
With the goal of creating general models that generalize compositionally in a large range of tasks, we show that several design decisions, such as position encodings, decoder type, weight sharing, model hyper-parameters, and formulation of the target task result in different inductive biases, with significant impact for compositional generalization\footnote{Source code: {\url{https://github.com/google-research/google-research/tree/master/compositional_transformers}}.}. 
We use a collection of twelve datasets designed to measure compositional generalization. In addition to six standard datasets commonly used in the literature (such as SCAN~\cite{lake2018generalization}, PCFG~\cite{hupkes2020compositionality}, CFQ~\cite{keysers2019measuring} and COGS~\cite{kim2020cogs}), we also use a set of basic algorithmic tasks (such as addition, duplication, or set intersection) that although not directly involving natural language,  are useful to obtain insights into what can and cannot be learned with different Transformer models. We also include tasks where we do not see significant improvements, to understand what types of compositional generalization are improved with our proposed modifications, and which are not.

The main contributions of this paper are: (1) A study of the Transformer design space, showing which design choices result in compositional learning biases across a variety of tasks. (2) state-of-the-art results in COGS, where we report a classification accuracy of 0.784 using an intermediate representation based on sequence tagging (compared to 0.35 for the best previously reported model~\cite{kim2020cogs}), and the productivity and systematicity splits of PCFG~\cite{hupkes2020compositionality}. 

The rest of this paper is organized as follows. Section~\ref{sec:background} provides some background on compositional generalization and Transformers. In Section~\ref{sec:datasets}, we present the datasets used in our empirical evaluation, which is presented in Section~\ref{sec:experiments}. The paper closes with a discussion on the implications of our results, and directions for future work.

\section{Background}\label{sec:background}

This section briefly provides background on compositional generalization and Transformer models.

\subsection{Compositional Generalization}\label{subsec:compositionality}

Compositional generalization can manifest in different ways. Hupkes et al.~(\citeyear{hupkes2020compositionality}) identified five different types, such as {\em systematicity} and {\em productivity} (extrapolation to longer sequences than those seen during training). %
{\em Systematicity} is the ability of recombining  known parts and rules in different ways than seen during training. The example in the introduction of knowing the meaning of ``jump``, ``jump twice`` and ``jax`` and from those inferring the meaning of ``jax twice`` is an example of systematicity. 
{\em Productivity}, on the other hand, is the ability to extrapolate to longer sequences than those seen during training. For example, consider the example of learning how to evaluate mathematical expressions of the form ``$3 + (4-(5*2))$''. An example of productivity would be to extrapolate to expressions with a larger number of parenthesis, or with deeper parenthesis nesting, than seen during training.
Hupkes et al.~(\citeyear{hupkes2020compositionality}) identify other forms of compositionality, such as {\em substitutivity}, {\em localism} or {\em overgeneralization}, but we will mostly focus on systematicity and productivity in this paper.

Compositional generalization is related to the general problem of {\em out-of-distribution generalization}. Hence, we can also see it as the problem of how models can discover {\em symmetries} in the domain (such as the existence of primitive operations or other regularities) that would generalize better to out-of-distribution samples than {\em shortcuts}~\cite{geirhos2020shortcut}, which would only work on the same distribution of examples seen during training. 

Early work focused on showing how different deep learning models do not generalize compositionally~\cite{livska2018memorize}. For example \citeauthor{livska2018memorize}~(\citeyear{livska2018memorize}) showed that while models like LSTMs are able to generalize compositionally, it is unlikely that gradient descent converges to a solution that does so (only about 2\% out of 50000 training runs achieved a generalization accuracy higher than 80\% in a compositional task, while they had almost perfect performance in training). 
Datasets like SCAN~\cite{lake2018generalization}, PCFG~\cite{hupkes2020compositionality}, Arithmetic language~\cite{veldhoen2016diagnostic}, or CFQ~\cite{keysers2019measuring} were proposed to show these effects. 


Work toward improving compositional generalization includes ideas like Syntactic attention~\cite{russin2019compositional}, 
increased pre-training~\cite{furrer2020compositional},
data augmentation~\cite{andreas2019good}, 
intermediate representations~\cite{herzig2021unlocking} or structure annotations~\cite{kim2021improving}. 
Specialized architectures that achieve good performance in specific compositional generalization tasks also exist.  For example, \citet{liu2020compositional} propose a model made up of a ``composer'' and a ``solver'', achieving perfect performance on SCAN. 
The most related concurrent work to ours is that of \citet{csordas2021devil}, who also 
showed gains in compositional generalization via relative attention. Additionally, in their work, they show that a key problem in some tasks is the {\em end of sequence} detection problem (when to stop producing output). Finally, they show that generalization accuracy keeps growing even when training accuracy maxes out, questioning early stopping approaches in compositional generalization. We note that training for longer might also improve our results, which we will explore in the future.

\subsection{Transformer Models}

\begin{figure*}[t!]
	\includegraphics[width=0.85\textwidth]{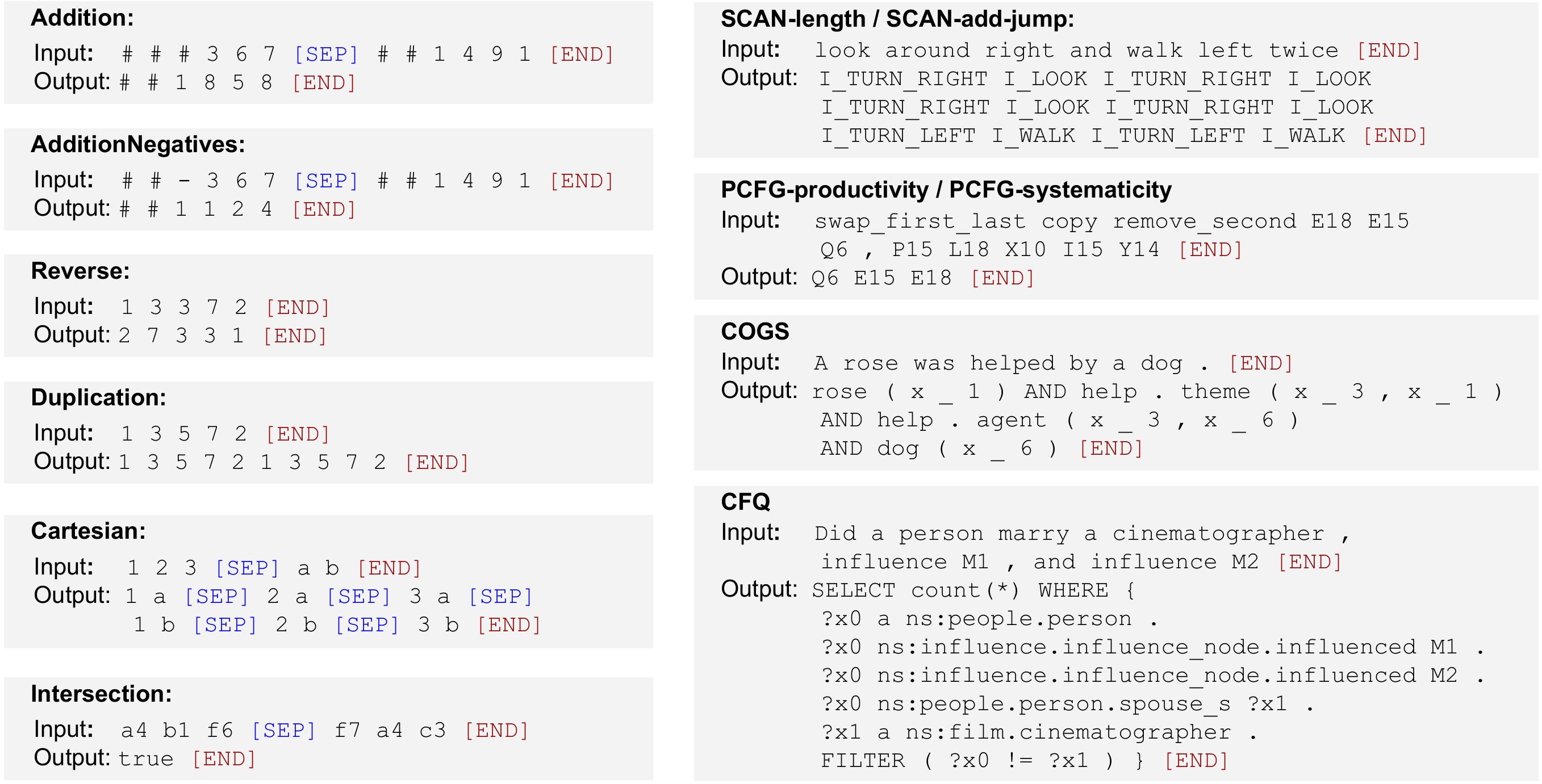}
	\centering
	\caption{Examples from the different datasets used in our experiments.}
	\label{fig:datasets}
\end{figure*}

Models based on Transformers~\cite{vaswani2017attention}, such as BERT~\cite{devlin2018bert}, or variants~\cite{yang2019xlnet,lan2019albert,raffel2019exploring} yield state-of-the-art results in many NLP tasks such as language modeling~\cite{child2019generating,sukhbaatar2019adaptive,rae2019compressive,kitaev2020reformer}, question answering~\cite{ainslie2020etc,lan2019albert,zaheer2020big,beltagy2020longformer}, and summarization~\cite{zhang2019hibert}. However, existing studies show that they do not have good compositional generalization. In this paper we will consider the original Transformer architecture and expand upon it. 

\begin{figure*}[t!]
	\includegraphics[width=0.85\textwidth]{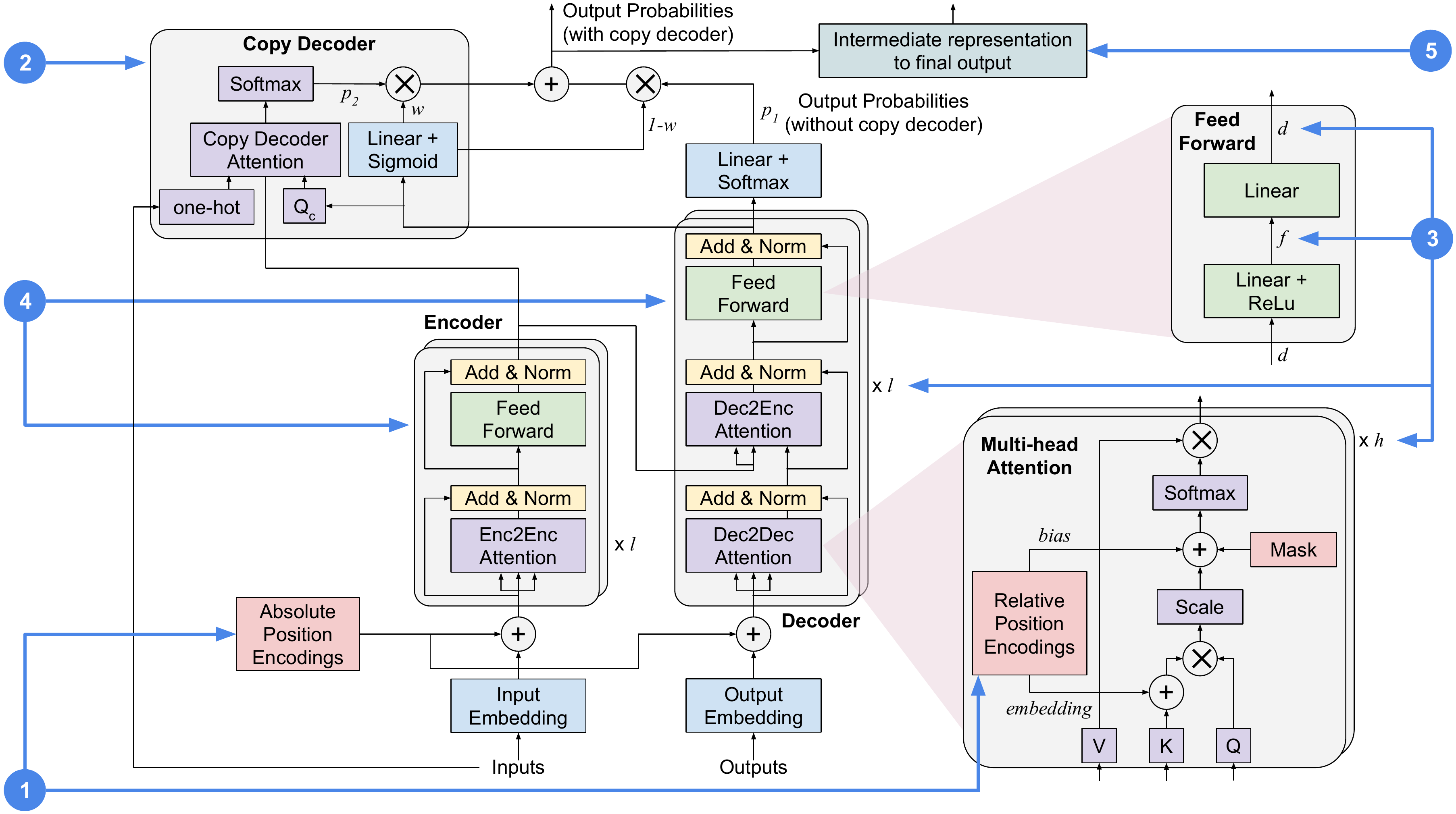}
	\centering
	\caption{An illustration of a Transformer, extended with the additional components necessary to explore the different dimensions we experiment with in this paper: {\bf (1)} position encodings, {\bf (2)} copy decoder, {\bf (3)} model size ($l, d, f, h$), {\bf (4)} weight sharing, 
	and {\bf (5)} intermediate representations.}
	\label{fig:dimensions}
\end{figure*}

The standard Transformer model  consists of two main components (see the center of Figure \ref{fig:dimensions}): an {\em encoder} and a {\em decoder}, each of which consists of a series of layers. Each layer contains an attention sublayer followed by a feed-forward sublayer (the decoder has two attention sublayers for decoder-to-decoder and decoder-to-encoder attention). The input of a Transformer is a sequence of token embeddings, and the output is a sequence of tokens generated one at a time by predicting based on the output distribution generated by the decoder. To provide a notion of token ``order'' a set of {\em position encodings} are typically added to the embedding of each input token to indicate sequence order. 

We will use $l$ to denote the number of encoder/decoder layers, 
$d$ for the dimensionality of token embeddings, $f$ for the intermediate dimensionality used by the feed-forward sublayer, and $h$ for the number of {\em attention-heads} in the attention sublayers. The original Transformer model used $l = 6$, $d = 512$, $f = 2048$ and $h = 8$, as their {\em base} configuration. In this paper, we use parameters much smaller than that, as we are evaluating the architectural decisions on relatively small datasets.

\section{Evaluation Datasets}\label{sec:datasets}

We use a collection of 12 datasets that require different types of compositional generalization. Six of those dataset consist of ``algorithmic'' tasks (addition, reversing lists, etc.), and six of them are standard datasets used to evaluate compositional generalization (most involving natural language). We note that our algorithmic tasks mostly require {\em productivity}-style compositional generalization, while other datasets also require {\em systematicity} or {\em synonimity}~\cite{hupkes2020compositionality}. Specifically, we used the following datasets (see Appendix E for details, and Figure \ref{fig:datasets} for examples):


    {\bf Addition} ({\em Add}): A synthetic addition task, where the input contains the digits of two integers, and the output should be the digits of their sum. 
    The training set contains numbers with up to 8 digits, and the test set contains numbers with 9 or 10 digits. Numbers are padded  to reach a length of 12.
    
    {\bf AdditionNegatives} ({\em AddNeg}): The same as the previous one, but 25\% of the numbers are negative (preceded with the \texttt{-} symbol).
    
    {\bf Reversing} ({\em Reverse}): Where the output is expected to be the input sequence in reverse order. 
    Training contains sequences of up to 16 digits, and the test set contains lengths between 17 to 24.
    
    {\bf Duplication} ({\em Dup}): The input is a sequence of digits and the output should be the same sequence, repeated twice. 
    Training contains sequences up to 16 digits, and test from 17 to 24.
    
    {\bf Cartesian} ({\em Cart}): The input contains two sequences of symbols, and the output should be their Cartesian product. 
    Training contains sequences of up to 6 symbols (7 or 8 for testing). 
    
    {\bf Intersection} ({\em Inters}): Given two sequences of symbols, 
    the output should be whether they have a non-empty intersection. Training contains sets with size 1 to 16, and testing 17 to 24.
    
    {\bf SCAN-length} ({\em SCAN-l}): The {\em length split} of the SCAN dataset~\cite{lake2018generalization}.
    
    {\bf SCAN-add-jump} ({\em SCAN-aj}): The {\em add primitive jump split} of the SCAN dataset~\cite{lake2018generalization}.
    
    {\bf PCFG-productivity} ({\em PCFG-p}): The productivity split of the PCFG dataset~\cite{hupkes2020compositionality}
    
    {\bf PCFG-sytematicity} ({\em PCFG-s}: The systematicity split of the PCFG dataset~\cite{hupkes2020compositionality}.
    
    {\bf COGS}: The generalization split of the COGS semantic parsing dataset~\cite{kim2020cogs}.
    
    {\bf CFQ-mcd1} ({\em CFQ}): The MCD1 split of the CFQ dataset~\cite{keysers2019measuring}.

Note that most of these datasets are trivial if the training and test sets come from the same distribution, and most Transformer models achieve near 100\% accuracy (except a few hard tasks like the Cartesian product or set intersection). Hence, splitting train and test data in a way that requires compositional generalization is key (e.g., having examples with larger sequences in the test set than in the training set). We want to make sure models do not just learn {\em shortcuts}~\cite{geirhos2020shortcut} that do not generalize to out-of-distribution data.

\section{Empirical Results}\label{sec:experiments}


In this section we present an evaluation of the compositional generalization abilities of Transformers with different architectural configurations. Specifically we evaluated: (1) the type of position encodings, (2) the use of copy decoders, (3) model size, (4) weight sharing, 
and (5) the use of intermediate representations for prediction (see Figure \ref{fig:dimensions}). 
For this systematic experimentation, we used small Transformer models, without pre-training (all models are trained from scratch). Even if previous work has reported benefits of pre-training in some compositional tasks (e.g., in CFQ~\cite{furrer2020compositional}), we aim at disentangling the effects of each architecture decision in and of itself, in the search for compositional inductive biases. 

Our results show that, while these decisions do not affect certain types of compositional generalization tasks, we see significant gains in others.

We report the average of at least 3 training runs (for algorithmic tasks, we use at least 5 training runs, and 10 for set intersection since they have a higher variance; see Appendix B). We use {\em sequence-level accuracy} as the evaluation metric: an output sequence with even just a single wrong token is considered wrong.

\subsection{Position Encodings}\label{subsec:encodings}

\begin{table*}[tb]\centering 
\resizebox{\textwidth}{!}{
\setlength{\tabcolsep}{4pt}
\begin{tabular}{l|cccccc|cccccc||c} 
& Add & AddNeg & Reverse & Dup & Cart & Inters & SCAN-l & SCAN-aj & PCFG-p & PCFG-s & COGS & CFQ & Avg. \\ \hline
{\em abs}  &  \makeDataBar{0.005}  &  \makeDataBar{0.042}  &  \makeDataBar{0.000}  &  \makeDataBar{0.000}  &  \makeDataBar{0.000}  &  \makeDataBar{0.500}  &  \makeDataBar{0.000}  &  \makeDataBar{0.003}  &  \makeDataBar{0.174}  &  \makeDataBar{0.434}  &  \makeDataBar{0.177}  &  \makeDataBar{0.304}  &  \makeAverageBar{0.137} \\ \hline
{\em rel-e}  &  \makeDataBar{0.004}  &  \makeDataBar{0.018}  &  \makeDataBar{0.422}  &  {\bf \makeDataBar{0.486}}  &  \makeDataBar{0.004}  &  \makeDataBar{0.501}  &  \makeDataBar{0.064}  &  \makeDataBar{0.003}  &  \makeDataBar{0.238}  &  \makeDataBar{0.451}  &  \makeDataBar{0.170}  &  {\bf \makeDataBar{0.322}}  &  \makeAverageBar{0.224} \\
{\em rel-b}  &  \makeDataBar{0.002}  &  \makeDataBar{0.005}  &  \makeDataBar{0.277}  &  \makeDataBar{0.362}  &  {\bf \makeDataBar{0.054}}  &  \makeDataBar{0.501}  &  \makeDataBar{0.049}  &  \makeDataBar{0.007}  &  \makeDataBar{0.042}  &  \makeDataBar{0.102}  &  \makeDataBar{0.126}  &  \makeDataBar{0.276}  &  \makeAverageBar{0.150} \\
{\em rel-eb}  &  \makeDataBar{0.003}  &  \makeDataBar{0.011}  &  \makeDataBar{0.486}  &  \makeDataBar{0.444}  &  \makeDataBar{0.000}  &  \makeDataBar{0.500}  &  \makeDataBar{0.089}  &  {\bf \makeDataBar{0.011}}  &  {\bf \makeDataBar{0.257}}  &  {\bf \makeDataBar{0.452}}  &  \makeDataBar{0.249}  &  \makeDataBar{0.290}  &  \makeAverageBar{0.233} \\ \hline
{\em rel2-e}  &  {\bf \makeDataBar{0.988}}  &  {\bf \makeDataBar{0.830}}  &  {\bf \makeDataBar{0.787}}  &  \makeDataBar{0.010}  &  \makeDataBar{0.000}  &  \makeDataBar{0.501}  &  \makeDataBar{0.032}  &  \makeDataBar{0.007}  &  \makeDataBar{0.159}  &  \makeDataBar{0.353}  &  {\bf \makeDataBar{0.259}}  &  \makeDataBar{0.322}  &  {\bf \makeAverageBar{0.354}} \\
{\em rel2-b}  &  \makeDataBar{0.140}  &  \makeDataBar{0.708}  &  \makeDataBar{0.056}  &  \makeDataBar{0.253}  &  \makeDataBar{0.000}  &  {\bf \makeDataBar{0.504}}  &  \makeDataBar{0.080}  &  \makeDataBar{0.002}  &  \makeDataBar{0.041}  &  \makeDataBar{0.117}  &  \makeDataBar{0.138}  &  \makeDataBar{0.319}  &  \makeAverageBar{0.197} \\
{\em rel2-eb}  &  \makeDataBar{0.978}  &  \makeDataBar{0.779}  &  \makeDataBar{0.737}  &  \makeDataBar{0.017}  &  \makeDataBar{0.000}  &  \makeDataBar{0.504}  &  {\bf \makeDataBar{0.091}}  &  \makeDataBar{0.010}  &  \makeDataBar{0.194}  &  \makeDataBar{0.374}  &  \makeDataBar{0.159}  &  \makeDataBar{0.311}  &  \makeAverageBar{0.346} \\
\end{tabular}		
}
\caption{Sequence-level accuracy for different position encoding methods. Bolded results represent the best results for each dataset in this table.}
\label{tbl:position} 
\end{table*}

While the original Transformer model~\cite{vaswani2017attention} and BERT~\cite{devlin2018bert} used {\em absolute position encodings}, later models such as T5~\cite{raffel2019exploring} or ETC~\cite{ainslie2020etc} use {\em relative position encodings}~\cite{shaw2018self}. 
Relative position encodings assign a {\em label} to each pair of tokens in the input (typically representing their relative distance in the input, up to a maximum radius). So, there is a label used for tokens attending to a token ``two positions to the right'', etc. 
One interesting thing about relative position encodings is that they are {\em position invariant}, i.e. two tokens that are $k$ positions apart will attend to each other in the same way, regardless of where they are in the sequence, and hence allowing models to capture further {\em symmetries} in the domain. 
%
We compare the following position encodings:

    {\bf abs}: sinusoidal absolute position encodings (as used in the original Transformer)\footnote{We did not experiment with learnable absolute position encodings, as test examples are longer than anything seen during training, hence containing untrained embeddings.}.
    
    {\bf rel-e}: relative position encodings, where the relative position label defines a learnable embedding that is added to the {\em key} during the attention process. We used a maximum local attention radius of 16, which means that we have the following relative position labels $\{l_{-16}, l_{-15}, ..., l_{-1}, l_{0}, l_{1}, ..., l_{15}, l_{16}\}$. Tokens that are further than 16 positions apart get the $l_{-16}$ or $l_{16}$ labels. 
    
    {\bf rel-b}: relative positions define a learnable bias that is added to the attention weight of each attention pair. This is the attention mechanism used by T5 (although they use a logarithmic scheme for representing relative positions).
    
    {\bf rel-eb}: relative position using both a learnable embedding vector and a learnable bias scalar.

While relative positions are straightforward for encoder-to-encoder and decoder-to-decoder attention, it is unclear what the relative positions should be for decoder-to-encoder. Hence, we tested three alternatives ({\bf rel2-e}, {\bf rel2-b} and {\bf rel2-eb} in our result tables). {\em rel-*} methods do not use relative position labels in decoder to encoder attention, and {\em rel2-*} do (where token $y_i$ in the decoder attending to token $x_j$ in the encoder will have label $l_{j-i}$.

\begin{table*}[tb]\centering 
\resizebox{\textwidth}{!}{
\setlength{\tabcolsep}{4pt}
\begin{tabular}{l|cccccc|cccccc||c} 
& Add & AddNeg & Reverse & Dup & Cart & Inters & SCAN-l & SCAN-aj & PCFG-p & PCFG-s & COGS & CFQ & Avg. \\ \hline
{\em abs}  &  \makeDataBar{0.005}  &  \makeDataBar{0.042}  &  \makeDataBar{0.000}  &  \makeDataBar{0.000}  &  \makeDataBar{0.000}  &  \makeDataBar{0.500}  &  \makeDataBar{0.000}  &  \makeDataBar{0.003}  &  \makeDataBar{0.174}  &  \makeDataBar{0.434}  &  \makeDataBar{0.177}  &  \makeDataBar{0.304}  &  \makeAverageBar{0.137} \\
{\em rel-eb}  &  \makeDataBar{0.003}  &  \makeDataBar{0.011}  &  \makeDataBar{0.486}  &  \makeDataBar{0.444}  &  \makeDataBar{0.000}  &  \makeDataBar{0.500}  &  \makeDataBar{0.089}  &  {\bf \makeDataBar{0.011}}  &  \makeDataBar{0.257}  &  \makeDataBar{0.452}  &  \makeDataBar{0.249}  &  \makeDataBar{0.290}  &  \makeAverageBar{0.233} \\
{\em rel2-eb}  &  {\bf \makeDataBar{0.978}}  &  \makeDataBar{0.779}  &  {\bf \makeDataBar{0.737}}  &  \makeDataBar{0.017}  &  \makeDataBar{0.000}  &  \makeDataBar{0.504}  &  {\bf \makeDataBar{0.091}}  &  \makeDataBar{0.010}  &  \makeDataBar{0.194}  &  \makeDataBar{0.374}  &  \makeDataBar{0.159}  &  {\bf \makeDataBar{0.311}}  &  \makeAverageBar{0.346} \\ \hline
{\em abs-c}  &  \makeDataBar{0.006}  &  \makeDataBar{0.021}  &  \makeDataBar{0.000}  &  \makeDataBar{0.000}  &  \makeDataBar{0.000}  &  \makeDataBar{0.501}  &  \makeDataBar{0.000}  &  \makeDataBar{0.003}  &  \makeDataBar{0.230}  &  \makeDataBar{0.390}  &  {\bf \makeDataBar{0.520}}  &  \makeDataBar{0.301}  &  \makeAverageBar{0.164} \\
{\em rel-eb-c}  &  \makeDataBar{0.004}  &  \makeDataBar{0.007}  &  \makeDataBar{0.271}  &  \makeDataBar{0.460}  &  \makeDataBar{0.000}  &  \makeDataBar{0.413}  &  \makeDataBar{0.026}  &  \makeDataBar{0.009}  &  {\bf \makeDataBar{0.342}}  &  \makeDataBar{0.541}  &  \makeDataBar{0.474}  &  \makeDataBar{0.311}  &  \makeAverageBar{0.238} \\
{\em rel2-eb-c}  &  \makeDataBar{0.977}  &  {\bf \makeDataBar{0.791}}  &  \makeDataBar{0.540}  &  \makeDataBar{0.283}  &  \makeDataBar{0.000}  &  {\bf \makeDataBar{0.528}}  &  \makeDataBar{0.043}  &  \makeDataBar{0.010}  &  \makeDataBar{0.336}  &  \makeDataBar{0.527}  &  \makeDataBar{0.511}  &  \makeDataBar{0.295}  &  {\bf \makeAverageBar{0.403}} \\
\end{tabular}
}
\caption{Sequence-level accuracy with and without copy decoding (models with a copy decoder are marked with a ``-c'' suffix). Bolded numbers are the best results for each dataset in this table.}
\label{tbl:decoder} 
\end{table*}

Table \ref{tbl:position} shows sequence-level classification accuracy for {\em small} Transformers ($l = 2$, $d = 64$, $f = 256$, $h = 4$). The right-most column shows the average accuracy across all datasets, and we can see that position encodings play a very significant role in the performance of the models. Going from 0.137 accuracy of the model with absolute position encodings up to 0.354 for a model with relative position encodings using embeddings (but no bias term), as well as relative positions for decoder-to-encoder attention. In general almost any type of relative position encodings help, but using embeddings helps more than using bias terms. Moreover, position encodings play a bigger role in algorithmic tasks. For example, in the {\em Add} and {\em AddNeg} tasks, models go from 0.005 and 0.042 accuracy to almost perfect accuracy (0.988 and 0.830 for the {\em rel2-e} model). Moreover tasks like SCAN or CFQ do not seem to be affected by position encodings, and 
using relative position encodings with only a bias term hurts in PCFG.


\subsection{Decoder Type}\label{subsec:decoder}

Many tasks (such as the {\em duplication} or {\em PCFG} datasets used in our experiments) require models able to learn things like ``output whatever is in position $k$ of the input'', rather than having to learn hard-coded rules for outputting the right token, depending on the input, a type of {\em symmetry} that can be captured with a {\em copy decoder}. 

The copy decoder in our experiments is fairly simple, and works as follows (Figure \ref{fig:dimensions}, top-left). It assumes that the input and output vocabularies are the same (we use the union of input and output vocabularies in our experiments). For a given token $x_i$ in the output (with final embedding $y_i$), in addition to the output probability distribution $p_1$ over the tokens in the vocabulary, the copy decoder produces a second distribution $p_2$, which is then mixed with $p_1$ via a weight $w$.  
$p_2$ is obtained by attending to the output of the last encoder layer (the attention {\em query} is calculated using a learnable weight matrix from $y_i$, the embeddings of the last encoder layer are used as the {\em keys}, and the {\em values} are a one-hot representation of the input tokens). The result is passed through a softmax layer, resulting in $p_2$.

\begin{table*}[tb]\centering 
\resizebox{\textwidth}{!}{
\setlength{\tabcolsep}{4pt}
\begin{tabular}{l|cccccc|cccccc||c} 
& Add & AddNeg & Reverse & Dup & Cart & Inters & SCAN-l & SCAN-aj & PCFG-p & PCFG-s & COGS & CFQ & Avg. \\ \hline
{\em small-2}  &  \makeDataBar{0.977}  &  \makeDataBar{0.791}  &  \makeDataBar{0.540}  &  \makeDataBar{0.283}  &  \makeDataBar{0.000}  &  {\bf \makeDataBar{0.528}}  &  \makeDataBar{0.043}  &  {\bf \makeDataBar{0.010}}  &  \makeDataBar{0.336}  &  \makeDataBar{0.527}  &  {\bf \makeDataBar{0.511}}  &  \makeDataBar{0.295}  &  \makeAverageBar{0.403} \\
{\em small-4}  &  \makeDataBar{0.986}  &  {\bf \makeDataBar{0.835}}  &  \makeDataBar{0.676}  &  {\bf \makeDataBar{0.572}}  &  \makeDataBar{0.000}  &  \makeDataBar{0.500}  &  \makeDataBar{0.170}  &  \makeDataBar{0.000}  &  \makeDataBar{0.499}  &  \makeDataBar{0.711}  &  \makeDataBar{0.501}  &  \makeDataBar{0.301}  &  {\bf \makeAverageBar{0.479}} \\
{\em small-6}  &  {\bf \makeDataBar{0.992}}  &  {\bf \makeDataBar{0.835}}  &  \makeDataBar{0.225}  &  \makeDataBar{0.000}  &  \makeDataBar{0.000}  &  \makeDataBar{0.203}  &  \makeDataBar{0.164}  &  \makeDataBar{0.002}  &  {\bf \makeDataBar{0.548}}  &  \makeDataBar{0.741}  &  \makeDataBar{0.476}  &  {\bf \makeDataBar{0.312}}  &  \makeAverageBar{0.375} \\ \hline
{\em large-2}  &  \makeDataBar{0.983}  &  \makeDataBar{0.811}  &  \makeDataBar{0.605}  &  \makeDataBar{0.503}  &  \makeDataBar{0.000}  &  \makeDataBar{0.500}  &  {\bf \makeDataBar{0.184}}  &  \makeDataBar{0.001}  &  \makeDataBar{0.535}  &  \makeDataBar{0.758}  &  \makeDataBar{0.498}  &  \makeDataBar{0.269}  &  \makeAverageBar{0.471} \\
{\em large-4}  &  \makeDataBar{0.957}  &  \makeDataBar{0.786}  &  {\bf \makeDataBar{0.684}}  &  \makeDataBar{0.523}  &  \makeDataBar{0.000}  &  \makeDataBar{0.400}  &  \makeDataBar{0.164}  &  \makeDataBar{0.004}  &  \makeDataBar{0.513}  &  {\bf \makeDataBar{0.770}}  &  \makeDataBar{0.462}  &  \makeDataBar{0.310}  &  \makeAverageBar{0.464} \\
{\em large-6}  &  \makeDataBar{0.978}  &  \makeDataBar{0.673}  &  \makeDataBar{0.423}  &  \makeDataBar{0.288}  &  \makeDataBar{0.000}  &  \makeDataBar{0.250}  &  \makeDataBar{0.144}  &  \makeDataBar{0.000}  &  \makeDataBar{0.530}  &  \makeDataBar{0.750}  &  \makeDataBar{0.451}  &  \makeDataBar{0.288}  &  \makeAverageBar{0.398} \\
\end{tabular}		
}
\caption{Sequence-level accuracy for models of different sizes. All models are variations of the {\em rel2-eb-c} model in Table \ref{tbl:decoder} ({\em small-2} is equivalent to {\em rel2-eb-c}). Bolded results represent the best results for each dataset in this table.}
\label{tbl:size} 
\end{table*}

Table \ref{tbl:decoder} shows sequence-level classification accuracy for models with and without a copy decoder. As can be seen in the last column ({\em Avg.}), having a copy decoder consistently helps performance, with all models using a copy decoder ({\em abs-c}, {\em rel-eb-c} and {\em rel2-eb-c}) outperforming their counterparts without a copy decoder. Moreover, we see that the copy decoder helps the most in {\em PCFG} and {\em COGS}, while it does not seem to help in some other tasks. 

Moreover, we would like to point out that there are other ways to set up copy decoders. For example Aky{\"u}rek et al. (\citeyear{akyurek2021lexicon}) propose defining a lexical translation layer in the copy decoder, which allows models to translate tokens in the input to tokens in the output (which is useful in tasks such as SCAN, which have disjoint vocabularies). In their work, they propose to initialize this layer via a lexicon learning task.

\subsection{Model Size}\label{subsec:size}

Next, we compare the effect of varying both the number of layers ($l$), as well as their size ($d$, $f$, $h$). Specifically, we tested models with number of layers $l$ equal to 2, 4 and 6, and layers of two sizes: {\em small} ($d = 64$, $f = 256$, $h = 4$), and {\em large} ($d = 128$, $f = 512$, $h = 8$). We denote these models {\em small-2}, {\em small-4}, {\em small-6}, {\em large-2}, {\em large-4}, and {\em large-6}. All of the models in this section are variants of {\em rel2-eb-c}, our previous best (see Appendix C for parameter counts of our models).

Table \ref{tbl:size} shows the sequence-level classification accuracy, showing a few interesting facts. First, in most algorithmic tasks, size does not help. Our hypothesis is that the logic required to learn these tasks does not require too many parameters, and large models probably overfit (e.g., like in {\em Duplication})\footnote{Further investigation showed that lowering the learning rate improves performance in the larger models, preventing the phenomenon seen in the {\em Duplication} dataset. Systematically exploring this is left for future work.}. Some datasets, however, do benefit from size. For example, most {\em large} models outperform their respective {\em small} ones in both variants of PCFG. These results are not unexpected, as most compositional generalization datasets contain idealized examples, often generated via some form of grammar, and have very small vocabularies (see Table \ref{tbl:datasets}). Hence, models might not benefit from size as much as on complex natural language tasks. 

\subsection{Weight Sharing}\label{subsec:sharing}

\begin{table*}[tb]\centering 
\resizebox{\textwidth}{!}{
\setlength{\tabcolsep}{4pt}
\begin{tabular}{l|cccccc|cccccc||c} 
& Add & AddNeg & Reverse & Dup & Cart & Inters & SCAN-l & SCAN-aj & PCFG-p & PCFG-s & COGS & CFQ & Avg. \\ \hline
{\em small-2s}  &  \makeDataBar{0.992}  &  \makeDataBar{0.809}  &  \makeDataBar{0.780}  &  \makeDataBar{0.750}  &  \makeDataBar{0.000}  &  {\bf \makeDataBar{0.699}}  &  \makeDataBar{0.022}  &  \makeDataBar{0.003}  &  \makeDataBar{0.313}  &  \makeDataBar{0.501}  &  \makeDataBar{0.450}  &  \makeDataBar{0.303}  &  \makeAverageBar{0.468} \\
{\em small-4s}  &  \makeDataBar{0.991}  &  \makeDataBar{0.955}  &  \makeDataBar{0.708}  &  \makeDataBar{0.580}  &  \makeDataBar{0.000}  &  \makeDataBar{0.500}  &  \makeDataBar{0.172}  &  {\bf \makeDataBar{0.017}}  &  \makeDataBar{0.534}  &  \makeDataBar{0.723}  &  \makeDataBar{0.445}  &  \makeDataBar{0.292}  &  \makeAverageBar{0.493} \\
{\em small-6s}  &  \makeDataBar{0.993}  &  \makeDataBar{0.933}  &  \makeDataBar{0.505}  &  \makeDataBar{0.000}  &  \makeDataBar{0.000}  &  \makeDataBar{0.500}  &  \makeDataBar{0.186}  &  \makeDataBar{0.000}  &  \makeDataBar{0.562}  &  \makeDataBar{0.780}  &  \makeDataBar{0.454}  &  \makeDataBar{0.295}  &  \makeAverageBar{0.434} \\ \hline
{\em large-2s}  &  {\bf \makeDataBar{0.997}}  &  \makeDataBar{0.894}  &  {\bf \makeDataBar{0.831}}  &  \makeDataBar{0.848}  &  \makeDataBar{0.000}  &  \makeDataBar{0.584}  &  \makeDataBar{0.033}  &  \makeDataBar{0.002}  &  \makeDataBar{0.511}  &  \makeDataBar{0.638}  &  \makeDataBar{0.465}  &  \makeDataBar{0.292}  &  \makeAverageBar{0.508} \\
{\em large-4s}  &  \makeDataBar{0.991}  &  \makeDataBar{0.915}  &  \makeDataBar{0.771}  &  {\bf \makeDataBar{0.882}}  &  \makeDataBar{0.000}  &  \makeDataBar{0.400}  &  \makeDataBar{0.186}  &  \makeDataBar{0.002}  &  \makeDataBar{0.589}  &  \makeDataBar{0.791}  &  {\bf \makeDataBar{0.475}}  &  {\bf \makeDataBar{0.327}}  &  {\bf \makeAverageBar{0.527}} \\
{\em large-6s}  &  \makeDataBar{0.985}  &  {\bf \makeDataBar{0.982}}  &  \makeDataBar{0.241}  &  \makeDataBar{0.000}  &  \makeDataBar{0.000}  &  \makeDataBar{0.500}  &  {\bf \makeDataBar{0.196}}  &  \makeDataBar{0.000}  &  {\bf \makeDataBar{0.634}}  &  {\bf \makeDataBar{0.828}}  &  \makeDataBar{0.454}  &  \makeDataBar{0.303}  &  \makeAverageBar{0.427} \\
\end{tabular}		
}
\caption{Sequence-level accuracy for all the models in Table \ref{tbl:size}, but sharing weights across layers.}
\label{tbl:sharing} 
\end{table*}

In this section we evaluate the effect of sharing weights across transformer layers. When weight sharing is activated, all learnable weights from all layers in the {\em encoder} are shared across layers, and the same is true across the layers of the {\em decoder}. 

Table \ref{tbl:sharing} shows the resulting performance of the models (to be compared with Table \ref{tbl:size}). Surprisingly, weight sharing significantly boosts compositional generalization accuracy, and almost all models achieve a higher average accuracy across all datasets than their equivalent models in Table~\ref{tbl:size}. In particular, datasets such as {\em AdditionNegatives} see a significant boost, with several models achieving higher than 0.9 accuracy (0.982 for {\em large-6s}). PCFG also significantly benefits from weight sharing, with the {\em large-6s} model achieving 0.634 and 0.828 in the productivity and systematicity versions, respectively. These are higher than previously reported results in the literature (using the original Transformer, which is a much larger model): 0.50 and 0.72~\cite{hupkes2020compositionality}. Notice, moreover that achieving good results in PCFG (or SCAN) is easy with specialized models. The important achievement is doing so with general purpose models. Our hypothesis is that a model with shared weights across layers might have a more suited inductive bias to learn primitive operations that are applied repeatedly to the input of the transformer (copying, reversing, duplicating, etc.).





\subsection{Intermediate Representations}\label{subsec:representation}

\begin{figure*}[t!]
	\includegraphics[width=0.9\textwidth]{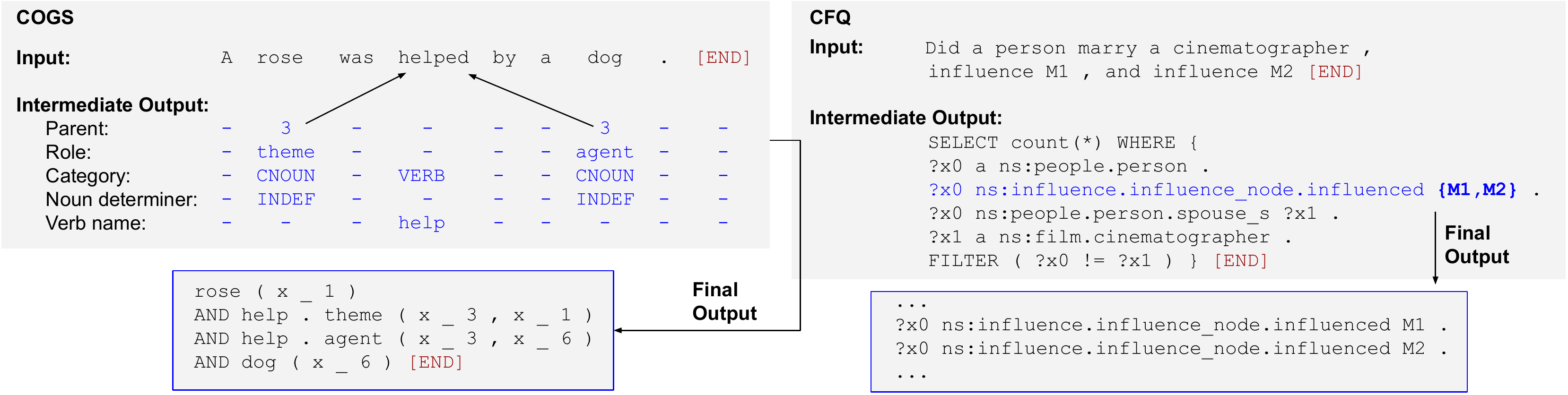}
	\centering
	\caption{Examples from the intermediate representations for COGs and CFQ. For COGs, we framed the task as {\em sequence tagging} and made the model predict 5 tags for each token; for CFQ we compressed Cartesian products.}
	\label{fig:datasets-intermediate}
\end{figure*}

The key idea of an {\em intermediate representation} is to define a different representation of the target output that is easier to generate by the model, but that can be easily mapped to the desired output. \citet{herzig2021unlocking} recently showed very promising results using this technique in several tasks. Defining useful intermediate representations is task-specific and not trivial. Thus we experimented with it in only two datasets: COGS and CFQ (Figure \ref{fig:datasets-intermediate}). 

\subsubsection{Intermediate Representation for COGS}

Our intermediate representation for COGs turns the task from seq2seq into a {\em sequence tagging} task. We ask the model to produce 5 tags for each input token: a {\em parent}, the {\em role} of the relation between the token and its parent (if applicable), the {\em category}, the {\em noun determiner} (for nouns) and the {\em verb name} (for verbs). With these tags, the original output can be constructed deterministically. One of the main advantages of this is that the model is naturally pushed to produce outputs with the correct length even for longer inputs (improving {\em productivity}).

For the sequence tagging formulation, we used only the encoder part of the Transformer and added five prediction heads, to predict each tag. For {\em role}, {\em category}, {\em noun determiner} and {\em verb name}, we simply had a dense layer with a Sigmoid activation function. For the  {\em parent} tag, we experimented with 3 different head types: {\em Absolute} used a dense layer with a Sigmoid activation to predict the absolute index of the parent in the input sequence ({\em -1} for no parent). {\em Relative} predicted the relative offset of the parent token with respect to the current token, or self for no parent. Finally, {\em Attention} used the attention weights from a new attention layer with 1 head to predict the parent.

\begin{table}[tb]\centering 
\resizebox{\columnwidth}{!}{
\setlength{\tabcolsep}{4pt}
\begin{tabular}{l|cc|cc} 
& \multicolumn{2}{c|}{seq2seq} & \multicolumn{2}{c}{tagging} \\ \hline
Model & abs & rel2-eb-c & abs & rel-eb \\
Size & small-2 & small-6s & small-2 & small-2s \\
Parent encoding &  &  & absolute & attention \\
\hline \hline \multicolumn{5}{l}{ Lexical Generalization: Primitives and Grammatical Roles } \\ \hline
Subject $\to$ Object (common noun) & \makeDataBar{0.309} & \makeDataBar{0.899} & \makeDataBar{0.911} & \makeDataBar{0.969} \\
Subject $\to$ Object (proper noun) & \makeDataBar{0.098} & \makeDataBar{0.429} & \makeDataBar{0.630} & \makeDataBar{0.826} \\
Object $\to$ Subject (common noun) & \makeDataBar{0.790} & \makeDataBar{0.936} & \makeDataBar{0.982} & \makeDataBar{0.978} \\
Object $\to$ Subject (proper noun) & \makeDataBar{0.207} & \makeDataBar{0.951} & \makeDataBar{0.993} & \makeDataBar{0.995} \\
Prim noun $\to$ Subject (common noun) & \makeDataBar{0.240} & \makeDataBar{0.913} & \makeDataBar{0.993} & \makeDataBar{0.988} \\
Prim noun $\to$ Subject (proper noun) & \makeDataBar{0.019} & \makeDataBar{0.772} & \makeDataBar{0.974} & \makeDataBar{0.996} \\
Prim noun $\to$ Object (common noun) & \makeDataBar{0.017} & \makeDataBar{0.902} & \makeDataBar{0.950} & \makeDataBar{0.953} \\
Prim noun $\to$ Object (proper noun) & \makeDataBar{0.000} & \makeDataBar{0.513} & \makeDataBar{0.651} & \makeDataBar{0.700} \\
Prim verb $\to$ Infinitival argument & \makeDataBar{0.000} & \makeDataBar{0.766} & \makeDataBar{0.000} & \makeDataBar{0.001} \\
\hline \hline  \multicolumn{5}{l}{ Lexical Generalization: Verb Argument Structure Alternation } \\ \hline
Active $\to$ Passive & \makeDataBar{0.604} & \makeDataBar{0.000} & \makeDataBar{0.697} & \makeDataBar{0.948} \\
Passive $\to$ Active & \makeDataBar{0.196} & \makeDataBar{0.001} & \makeDataBar{0.535} & \makeDataBar{0.897} \\
Object-omitted transitive $\to$ Transitive & \makeDataBar{0.275} & \makeDataBar{0.003} & \makeDataBar{0.527} & \makeDataBar{0.926} \\
Unaccusative $\to$ Transitive & \makeDataBar{0.069} & \makeDataBar{0.003} & \makeDataBar{0.528} & \makeDataBar{0.787} \\
Double object dative $\to$ PP dative & \makeDataBar{0.819} & \makeDataBar{0.000} & \makeDataBar{0.590} & \makeDataBar{0.958} \\
PP dative $\to$ Double object dative & \makeDataBar{0.404} & \makeDataBar{0.004} & \makeDataBar{0.771} & \makeDataBar{0.850} \\
\hline \hline \multicolumn{5}{l}{ Lexical Generalization: Verb Class } \\ \hline
Agent NP $\to$ Unaccusative Subject & \makeDataBar{0.399} & \makeDataBar{0.951} & \makeDataBar{0.784} & \makeDataBar{1.000} \\
Theme NP $\to$ Obj-omitted trans Subj & \makeDataBar{0.688} & \makeDataBar{0.965} & \makeDataBar{0.791} & \makeDataBar{0.701} \\
Theme NP $\to$ Unergative subject & \makeDataBar{0.694} & \makeDataBar{0.966} & \makeDataBar{0.930} & \makeDataBar{0.771} \\
\hline \hline
\multicolumn{5}{l}{ Structural Generalization: Phrases and Grammatical Roles } \\ \hline
Obj-mod PP $\to$ Subj-mod PP & \makeDataBar{0.000} & \makeDataBar{0.000} & \makeDataBar{0.000} & \makeDataBar{0.299} \\
\hline \hline \multicolumn{5}{l}{ Structural Generalization: Deeper Recursion } \\ \hline
Depth generalization: PP modifiers & \makeDataBar{0.003} & \makeDataBar{0.000} & \makeDataBar{0.138} & \makeDataBar{0.681} \\
Depth generalization: Sentential comp & \makeDataBar{0.000} & \makeDataBar{0.000} & \makeDataBar{0.000} & \makeDataBar{0.233} \\ \hline \hline
Overall & \makeDataBar{0.278} & \makeDataBar{0.475} & \makeDataBar{0.637} & \makeDataBar{0.784} \\
\end{tabular}		
}
\caption{Sequence-level accuracy in different generalization subsets in COGS for both seq2seq and sequence tagging models. PP stands for prepositional phrase.}
\label{tbl:cogs} 
\end{table}

Table \ref{tbl:cogs} shows the experimental results comparing a few configurations of this new tagging approach to a few configurations of the seq2seq approach (see Appendix D for all other configurations).
Examples in the structural generalization tasks are typically longer than in the training set and require {\em productivity}. All the models tested in the original COGS paper~\cite{kim2020cogs} (and all of our seq2seq approaches above) achieved 0 accuracy in this category. The {\em small-6s} seq2seq model improves the overall performance from 0.278 to 0.475, but curiously has near 0 performance on {\em Verb Argument Structure Alternation} tasks, worse than the base {\em abs} model. 

The intermediate representation based on tagging works much better. The base {\em abs} tagging model manages to get non-zero performance on one structural generalization task, which suggests that enforcing the right output length helps. Finally, when predicting the parent directly from attention weights, the structural generalization tasks score 0.2-0.7, compared to our previous near 0 scores (see Appendix D for common types of errors).

Overall, the sequence tagging intermediate representation achieves a much higher accuracy, with one model reaching 0.784, higher than any previously reported performance in COGS in the literature, to the best of our knowledge. This suggests that the encoder has the power to parse the input correctly, but maybe the decoder is not capable of generating the correct output sequence from the encoder in the full transformer.

\subsubsection{Intermediate Representation for CFQ}

One of the difficulties in the CFQ dataset is that models need to learn to perform Cartesian products (e.g., for questions like ``who directed and acted in M1 and M2?'', the model needs to expand to ``directed M1'', ``directed M2'', ``acted in M1'' and ``acted in M2''). However, as shown in our experiments above, this is a very hard task to learn. 
Hence, we followed the same idea as in Herzig et al.~(\citeyear{herzig2021unlocking}), and defined an intermediate representation that removes the need to learn Cartesian products by allowing triples of the form {\em (entity list) - (relation list) - (entity list)}. 

\begin{table}[tb]\centering 
\setlength{\tabcolsep}{4pt}
\begin{small}
\begin{tabular}{l|c|c} 
& CFQ & CFQ-im \\ \hline
{\em abs}  &  \makeDataBar{0.304}  &  \makeDataBar{0.541}  \\ \hline
{\em rel-eb}  &  \makeDataBar{0.290}  &  {\bf \makeDataBar{0.555}} \\
{\em rel2-eb}  &  \makeDataBar{0.311}  &  \makeDataBar{0.541}  \\
{\em rel-eb-c}  &  \makeDataBar{0.311}  &  \makeDataBar{0.541}  \\
{\em rel2-eb-c}  &  \makeDataBar{0.295}  &  \makeDataBar{0.519}  \\
{\em large-4}  &  \makeDataBar{0.310}  &  \makeDataBar{0.541}  \\
{\em large-4s}  &  {\bf \makeDataBar{0.327}}  &  \makeDataBar{0.474}  \\
\end{tabular}		
\end{small}
\caption{Sequence-level accuracy for different models for the original CFQ, and for CFQ with intermediate representations ({\em CFQ-im}). The top 5 models are small models with 2 layers, and the last four models are variants of {\em rel2-eb-c} (used in Tables \ref{tbl:size} and \ref{tbl:sharing}).}
\label{tbl:cfq-intermediate} 
\end{table}

Table \ref{tbl:cfq-intermediate} shows the sequence-level classification accuracy for models on CFQ and on the version with intermediate representations ({\em CFQ-im}). While the different variations on Transformer models have little affect on the performance, the use of an intermediate representation significantly improves performance, going from around 0.3 accuracy for most Transformer models to over 0.5, and up to 0.555 for the {\em rel-eb} model. This is consistent with the results reported by Herzig et al.~(\citeyear{herzig2021unlocking}).

\section{Discussion}\label{sec:discussion}

An overall trend is that algorithmic tasks seem to be greatly affected by the different architecture design decisions we explored. In all datasets, except for Cartesian product, there is at least one combination in our experiments that achieved high performance (close to 0.8 accuracy or higher). 
Cartesian products remain an open challenge for future work, where one of the big obstacles is learning to produce much longer outputs than seen during training (output is quadratic with respect to input size).

There are some datasets, such as {\em SCAN-aj}, where we did not see large improvements in performance. The main obstacle is learning to handle a symbol (``jump'') having seen it very few times (or even just once) during training (this also happens in some types of generalization in COGS). None of the variations we experimented with were enough to handle this type of compositionality either.




In conclusion, we observed: 

\begin{enumerate}
\item {\em relative position encodings} (when both embeddings and biases are used) seem to never be detrimental (they either provided gains, or did not affect). Results indicate this significantly helps in {\em productivity}. Moreover, for tasks where positional information is important (such as addition, or reversing), adding positional encodings to decoder2encoder attention provided significant benefits. Finally, as Table \ref{tbl:position} shows, for relative position embeddings to be beneficial, using embeddings was necessary; only using relative position biases was not enough.

\item Adding a {\em copy decoder} was generally beneficial. We saw some occasional degradation in some tasks (e.g., {\em Reverse}), but these are high variance tasks (see Table \ref{tbl:all-stddev} in the Appendix), where results are more uncertain.

\item {\em Model size} in terms of embedding dimensions, helped generally. Going from 2 to 4 layers provided a slight benefit in general. Our experiments show going to 6 layers hurt performance, but as noted earlier, additional (unreported preliminary) experiments indicated larger models might need smaller learning rates, with which they also seem to improve performance (systematic exploration of this is future work).

\item {\em Weight sharing} seems to benefit in tasks where there are a clear set of primitives that have to be learned (PCFG in particular), or algorithmic tasks, but it seems to hurt in COGs. Hence, weight sharing does not provide general benefits as the previous modifications.

\item Intermediate representations, although dataset-specific, significantly help when they can be defined, as expected. 
\end{enumerate}

\section{Conclusions}\label{sec:conclusions}

This paper presented an empirical study of the design space of Transformer models, evaluated in a collection of benchmarks for compositional generalization in language and algorithmic tasks. Our results show that, compared to a baseline Transformer, significant gains in compositional generalization can be achieved. 
Specifically, the baseline Transformer achieved an average sequence-level accuracy of 0.137, while we showed this can increase to up to 0.527 with some design changes. Accuracy levels of up to 0.493 can be achieved without increasing the parameter count of our baseline model (see Appendix C for parameter counts). 
Moreover, we achieved state-of-the-art results in COGS (at the time of submission), showing 0.784 accuracy on the generalization set, and two PCFG splits (0.634 and 0.828 respectively). This shows that a key factor in training models that generalize compositionally is to provide the right inductive biases. 

As part of our future work, we want to explore more dimensions, such as pre-training and optimizer parameters, and study the implications of our results in compositional generalization in large models on real world tasks.


\bibliographystyle{acl_natbib}

\clearpage

\appendix

\section{Implementation Details}

We used a standard Transformer implementation\footnote{\url{https://www.tensorflow.org/tutorials/text/transformer}}, and added all the proposed variations on top of it. All experiments were run on machines with a single CPU and a single Tesla V100 GPU. All parameters were left to their default values from the original implementation, including the learning rate schedule (which could probably be further tweaked if state-of-the-art results are sought), as we were just aiming to compare inductive biases, rather than aim for SOTA results.

Additionally, we would like to highlight some implementation details, which surprisingly had large effects on our experimental results. {\em Layer normalization} operations in our Transformer implementation were done {\em after} each sublayer (attention and feed forward). Embedding layers were initialized with the Keras default ``uniform'' Keras initializer (uniform random distribution in the range $[-0.05, 0.05]$). 
Dense layers were initialized also with the Keras default Glorot initializer (uniform random distribution with mean 0 and standard deviation $\sqrt{2/(fan\_in+fan\_out)}$) \cite{glorot2010}.
While these details might not seem that important, we were unable to reproduce some of the results reported above using a re-implementation of the Transformer model in Flax, which used different defaults (and layer normalization before each sublayer rather than after) unless we changed these implementation details to match those of the Keras implementation. This indicates that these low-level details also have an effect on the learning bias of the models, with an impact in compositional generalization, which we plan to study in the future.

\begin{table}[b]\centering 
\resizebox{\columnwidth}{!}{
\begin{tabular}{l|c|c|c|c} 
{\em Dataset} & $|\mathit{Train}|$ & $|\mathit{Test}|$ & $|\mathit{Vocab}|$ & {\em  Epochs} \\ \hline
Add     & 200000 & 1024 & 14 & 2 \\
AddNeg  & 200000 & 1024 & 16 & 10 \\
Reverse & 200000 & 1024 & 14 & 2 \\
Dup     & 200000 & 1024 & 14 & 4 \\
Cart    & 200000 & 1024 & 24 & 4 \\
Inters  & 200000 & 1024 & 106 & 8 \\
SCAN-l  & 16989 & 3919 & 25 & 24 \\
SCAN-aj & 14669 & 7705 & 25 & 24 \\
PCFG-p  & 81011 & 11331 & 537 & 20 \\
PCFG-s  & 82167 & 10175 & 537 & 20 \\
COGS    & 24155 & 21000 & 876 & 16 \\
CFQ     & 95743 & 11968 & 184 & 16 \\
\end{tabular}		
}
\caption{Size of the training/test sets, vocab and training epochs we used for the different datasets.}
\label{tbl:datasets} 
\end{table}

\section{Detailed Results}

Table \ref{tbl:all-avg} shows the average sequence-level accuracy for all the models evaluated in this paper, all in one table. We used the same names as used in the paper (as models {\em rel2-eb-c} and {\em small-2} both refer to the same model, we included the row twice, with both names, for clarity). 

Table \ref{tbl:all-max} shows the maximum accuracy each model achieved in each dataset out of the 3 to 10 repetitions we did for each dataset. Recall we used 3 repetitions for SCAN-l, SCAN-aj, PCFG-p, PCFG-s, COGS and CFQ, 5 repetitions for Add, AddNeg, Reverse, Dup and Cart, and 10 repetitions for Inters (as it was the dataset where we saw more extreme results). An interesting phenomenon observed in the {\em Inters} dataset is that models tend to achieve either random accuracy (around 0.5), or perfect accuracy (1.0). Very rarely models achieve intermediate values. This support the {\em needle-in-a-haystack} argument of \citeauthor{livska2018memorize}~(\citeyear{livska2018memorize}), who saw that while LSTMs have the capability of generalize compositionally, what happens in practice is that gradient descent has a very low probability of converging to weights that do so (finding the ``compositional needle'' in a haystack). We observed a similar thing in our experiments, but saw that some Transformer architectures resulted in an increased chance of finding this {\em needle}.

Table \ref{tbl:all-stddev} shows the standard deviation in the sequence-level accuracy we observed in our experiments. As can be seen, the algorithmic tasks result in a much larger standard deviation. In some datasets (e.g., Add and Inters) it was common for morels to either achieve near 0\% accuracy (50\% in Inters) or near 100\% accuracy, but few values in between.

\begin{table*}[tb]\centering 
\resizebox{\textwidth}{!}{
\setlength{\tabcolsep}{4pt}
\begin{tabular}{l|cccccc|cccccc||c} 
& Add & AddNeg & Reverse & Dup & Cart & Inters & SCAN-l & SCAN-aj & PCFG-p & PCFG-s & COGS & CFQ & Avg. \\ \hline
{\em abs}  &  \makeDataBar{0.005}  &  \makeDataBar{0.042}  &  \makeDataBar{0.000}  &  \makeDataBar{0.000}  &  \makeDataBar{0.000}  &  \makeDataBar{0.500}  &  \makeDataBar{0.000}  &  \makeDataBar{0.003}  &  \makeDataBar{0.174}  &  \makeDataBar{0.434}  &  \makeDataBar{0.177}  &  \makeDataBar{0.304}  &  \makeAverageBar{0.137} \\ \hline
{\em rel-e}  &  \makeDataBar{0.004}  &  \makeDataBar{0.018}  &  \makeDataBar{0.422}  &  \makeDataBar{0.486}  &  \makeDataBar{0.004}  &  \makeDataBar{0.501}  &  \makeDataBar{0.064}  &  \makeDataBar{0.003}  &  \makeDataBar{0.238}  &  \makeDataBar{0.451}  &  \makeDataBar{0.170}  &  \makeDataBar{0.322}  &  \makeAverageBar{0.224} \\
{\em rel-b}  &  \makeDataBar{0.002}  &  \makeDataBar{0.005}  &  \makeDataBar{0.277}  &  \makeDataBar{0.362}  &  {\bf \makeDataBar{0.054}}  &  \makeDataBar{0.501}  &  \makeDataBar{0.049}  &  \makeDataBar{0.007}  &  \makeDataBar{0.042}  &  \makeDataBar{0.102}  &  \makeDataBar{0.126}  &  \makeDataBar{0.276}  &  \makeAverageBar{0.150} \\
{\em rel-eb}  &  \makeDataBar{0.003}  &  \makeDataBar{0.011}  &  \makeDataBar{0.486}  &  \makeDataBar{0.444}  &  \makeDataBar{0.000}  &  \makeDataBar{0.500}  &  \makeDataBar{0.089}  &  \makeDataBar{0.011}  &  \makeDataBar{0.257}  &  \makeDataBar{0.452}  &  \makeDataBar{0.249}  &  \makeDataBar{0.290}  &  \makeAverageBar{0.233} \\ \hline
{\em rel2-e}  &  \makeDataBar{0.988}  &  \makeDataBar{0.830}  &  \makeDataBar{0.787}  &  \makeDataBar{0.010}  &  \makeDataBar{0.000}  &  \makeDataBar{0.501}  &  \makeDataBar{0.032}  &  \makeDataBar{0.007}  &  \makeDataBar{0.159}  &  \makeDataBar{0.353}  &  \makeDataBar{0.259}  &  \makeDataBar{0.322}  &  \makeAverageBar{0.354} \\
{\em rel2-b}  &  \makeDataBar{0.140}  &  \makeDataBar{0.708}  &  \makeDataBar{0.056}  &  \makeDataBar{0.253}  &  \makeDataBar{0.000}  &  \makeDataBar{0.504}  &  \makeDataBar{0.080}  &  \makeDataBar{0.002}  &  \makeDataBar{0.041}  &  \makeDataBar{0.117}  &  \makeDataBar{0.138}  &  \makeDataBar{0.319}  &  \makeAverageBar{0.197} \\
{\em rel2-eb}  &  \makeDataBar{0.978}  &  \makeDataBar{0.779}  &  \makeDataBar{0.737}  &  \makeDataBar{0.017}  &  \makeDataBar{0.000}  &  \makeDataBar{0.504}  &  \makeDataBar{0.091}  &  \makeDataBar{0.010}  &  \makeDataBar{0.194}  &  \makeDataBar{0.374}  &  \makeDataBar{0.159}  &  \makeDataBar{0.311}  &  \makeAverageBar{0.346} \\ \hline
{\em abs-c}  &  \makeDataBar{0.006}  &  \makeDataBar{0.021}  &  \makeDataBar{0.000}  &  \makeDataBar{0.000}  &  \makeDataBar{0.000}  &  \makeDataBar{0.501}  &  \makeDataBar{0.000}  &  \makeDataBar{0.003}  &  \makeDataBar{0.230}  &  \makeDataBar{0.390}  &  \makeDataBar{0.520}  &  \makeDataBar{0.301}  &  \makeAverageBar{0.164} \\
{\em rel-eb-c}  &  \makeDataBar{0.004}  &  \makeDataBar{0.007}  &  \makeDataBar{0.271}  &  \makeDataBar{0.460}  &  \makeDataBar{0.000}  &  \makeDataBar{0.413}  &  \makeDataBar{0.026}  &  \makeDataBar{0.009}  &  \makeDataBar{0.342}  &  \makeDataBar{0.541}  &  \makeDataBar{0.474}  &  \makeDataBar{0.311}  &  \makeAverageBar{0.238} \\
{\em rel2-eb-c}  &  \makeDataBar{0.977}  &  \makeDataBar{0.791}  &  \makeDataBar{0.540}  &  \makeDataBar{0.283}  &  \makeDataBar{0.000}  &  \makeDataBar{0.528}  &  \makeDataBar{0.043}  &  \makeDataBar{0.010}  &  \makeDataBar{0.336}  &  \makeDataBar{0.527}  &  {\bf \makeDataBar{0.511}}  &  \makeDataBar{0.295}  &  \makeAverageBar{0.403} \\ \hline
{\em small-2}  &  \makeDataBar{0.977}  &  \makeDataBar{0.791}  &  \makeDataBar{0.540}  &  \makeDataBar{0.283}  &  \makeDataBar{0.000}  &  \makeDataBar{0.528}  &  \makeDataBar{0.043}  &  \makeDataBar{0.010}  &  \makeDataBar{0.336}  &  \makeDataBar{0.527}  &  {\bf \makeDataBar{0.511}}  &  \makeDataBar{0.295}  &  \makeAverageBar{0.403} \\
{\em small-4}  &  \makeDataBar{0.986}  &  \makeDataBar{0.835}  &  \makeDataBar{0.676}  &  \makeDataBar{0.572}  &  \makeDataBar{0.000}  &  \makeDataBar{0.500}  &  \makeDataBar{0.170}  &  \makeDataBar{0.000}  &  \makeDataBar{0.499}  &  \makeDataBar{0.711}  &  \makeDataBar{0.501}  &  \makeDataBar{0.301}  &  \makeAverageBar{0.479} \\
{\em small-6}  &  \makeDataBar{0.992}  &  \makeDataBar{0.835}  &  \makeDataBar{0.225}  &  \makeDataBar{0.000}  &  \makeDataBar{0.000}  &  \makeDataBar{0.203}  &  \makeDataBar{0.164}  &  \makeDataBar{0.002}  &  \makeDataBar{0.548}  &  \makeDataBar{0.741}  &  \makeDataBar{0.476}  &  \makeDataBar{0.312}  &  \makeAverageBar{0.375} \\ \hline
{\em large-2}  &  \makeDataBar{0.983}  &  \makeDataBar{0.811}  &  \makeDataBar{0.605}  &  \makeDataBar{0.503}  &  \makeDataBar{0.000}  &  \makeDataBar{0.500}  &  \makeDataBar{0.184}  &  \makeDataBar{0.001}  &  \makeDataBar{0.535}  &  \makeDataBar{0.758}  &  \makeDataBar{0.498}  &  \makeDataBar{0.269}  &  \makeAverageBar{0.471} \\
{\em large-4}  &  \makeDataBar{0.957}  &  \makeDataBar{0.786}  &  \makeDataBar{0.684}  &  \makeDataBar{0.523}  &  \makeDataBar{0.000}  &  \makeDataBar{0.400}  &  \makeDataBar{0.164}  &  \makeDataBar{0.004}  &  \makeDataBar{0.513}  &  \makeDataBar{0.770}  &  \makeDataBar{0.462}  &  \makeDataBar{0.310}  &  \makeAverageBar{0.464} \\
{\em large-6}  &  \makeDataBar{0.978}  &  \makeDataBar{0.673}  &  \makeDataBar{0.423}  &  \makeDataBar{0.288}  &  \makeDataBar{0.000}  &  \makeDataBar{0.250}  &  \makeDataBar{0.144}  &  \makeDataBar{0.000}  &  \makeDataBar{0.530}  &  \makeDataBar{0.750}  &  \makeDataBar{0.451}  &  \makeDataBar{0.288}  &  \makeAverageBar{0.398} \\ \hline
{\em small-2s}  &  \makeDataBar{0.992}  &  \makeDataBar{0.809}  &  \makeDataBar{0.780}  &  \makeDataBar{0.750}  &  \makeDataBar{0.000}  &  {\bf \makeDataBar{0.699}}  &  \makeDataBar{0.022}  &  \makeDataBar{0.003}  &  \makeDataBar{0.313}  &  \makeDataBar{0.501}  &  \makeDataBar{0.450}  &  \makeDataBar{0.303}  &  \makeAverageBar{0.468} \\
{\em small-4s}  &  \makeDataBar{0.991}  &  \makeDataBar{0.955}  &  \makeDataBar{0.708}  &  \makeDataBar{0.580}  &  \makeDataBar{0.000}  &  \makeDataBar{0.500}  &  \makeDataBar{0.172}  &  {\bf \makeDataBar{0.017}}  &  \makeDataBar{0.534}  &  \makeDataBar{0.723}  &  \makeDataBar{0.445}  &  \makeDataBar{0.292}  &  \makeAverageBar{0.493} \\
{\em small-6s}  &  \makeDataBar{0.993}  &  \makeDataBar{0.933}  &  \makeDataBar{0.505}  &  \makeDataBar{0.000}  &  \makeDataBar{0.000}  &  \makeDataBar{0.500}  &  \makeDataBar{0.186}  &  \makeDataBar{0.000}  &  \makeDataBar{0.562}  &  \makeDataBar{0.780}  &  \makeDataBar{0.454}  &  \makeDataBar{0.295}  &  \makeAverageBar{0.434} \\ \hline
{\em large-2s}  &  {\bf \makeDataBar{0.997}}  &  \makeDataBar{0.894}  &  {\bf \makeDataBar{0.831}}  &  \makeDataBar{0.848}  &  \makeDataBar{0.000}  &  \makeDataBar{0.584}  &  \makeDataBar{0.033}  &  \makeDataBar{0.002}  &  \makeDataBar{0.511}  &  \makeDataBar{0.638}  &  \makeDataBar{0.465}  &  \makeDataBar{0.292}  &  \makeAverageBar{0.508} \\
{\em large-4s}  &  \makeDataBar{0.991}  &  \makeDataBar{0.915}  &  \makeDataBar{0.771}  &  {\bf \makeDataBar{0.882}}  &  \makeDataBar{0.000}  &  \makeDataBar{0.400}  &  \makeDataBar{0.186}  &  \makeDataBar{0.002}  &  \makeDataBar{0.589}  &  \makeDataBar{0.791}  &  \makeDataBar{0.475}  &  {\bf \makeDataBar{0.327}}  &  {\bf \makeAverageBar{0.527}} \\
{\em large-6s}  &  \makeDataBar{0.985}  &  {\bf \makeDataBar{0.982}}  &  \makeDataBar{0.241}  &  \makeDataBar{0.000}  &  \makeDataBar{0.000}  &  \makeDataBar{0.500}  &  {\bf \makeDataBar{0.196}}  &  \makeDataBar{0.000}  &  {\bf \makeDataBar{0.634}}  &  {\bf \makeDataBar{0.828}}  &  \makeDataBar{0.454}  &  \makeDataBar{0.303}  &  \makeAverageBar{0.427} \\
\end{tabular}		
}
\caption{Average sequence-level accuracy for all the models evaluated in this paper.}
\label{tbl:all-avg} 
\end{table*}

\begin{table*}[tb]\centering 
\resizebox{\textwidth}{!}{
\setlength{\tabcolsep}{4pt}
\begin{tabular}{l|cccccc|cccccc} 
& Add & AddNeg & Reverse & Dup & Cart & Inters & SCAN-l & SCAN-aj & PCFG-p & PCFG-s & COGS & CFQ \\ \hline
{\em abs}  &  \makeDataBar{0.008}  &  \makeDataBar{0.131}  &  \makeDataBar{0.002}  &  \makeDataBar{0.000}  &  \makeDataBar{0.000}  &  \makeDataBar{0.500}  &  \makeDataBar{0.000}  &  \makeDataBar{0.008}  &  \makeDataBar{0.191}  &  \makeDataBar{0.462}  &  \makeDataBar{0.211}  &  \makeDataBar{0.326}  \\ \hline
{\em rel-e}  &  \makeDataBar{0.010}  &  \makeDataBar{0.059}  &  \makeDataBar{0.597}  &  \makeDataBar{0.908}  &  \makeDataBar{0.034}  &  \makeDataBar{0.511}  &  \makeDataBar{0.115}  &  \makeDataBar{0.007}  &  \makeDataBar{0.257}  &  \makeDataBar{0.496}  &  \makeDataBar{0.281}  &  \makeDataBar{0.346}  \\
{\em rel-b}  &  \makeDataBar{0.004}  &  \makeDataBar{0.016}  &  \makeDataBar{0.331}  &  \makeDataBar{0.417}  &  {\bf \makeDataBar{0.137}}  &  \makeDataBar{0.510}  &  \makeDataBar{0.072}  &  \makeDataBar{0.013}  &  \makeDataBar{0.047}  &  \makeDataBar{0.112}  &  \makeDataBar{0.170}  &  \makeDataBar{0.305}  \\
{\em rel-eb}  &  \makeDataBar{0.006}  &  \makeDataBar{0.018}  &  \makeDataBar{0.658}  &  \makeDataBar{0.795}  &  \makeDataBar{0.001}  &  \makeDataBar{0.502}  &  \makeDataBar{0.129}  &  \makeDataBar{0.023}  &  \makeDataBar{0.268}  &  \makeDataBar{0.528}  &  \makeDataBar{0.306}  &  \makeDataBar{0.333}  \\ \hline
{\em rel2-e}  &  {\bf \makeDataBar{1.000}}  &  \makeDataBar{0.943}  &  \makeDataBar{0.917}  &  \makeDataBar{0.038}  &  \makeDataBar{0.000}  &  \makeDataBar{0.512}  &  \makeDataBar{0.058}  &  \makeDataBar{0.018}  &  \makeDataBar{0.182}  &  \makeDataBar{0.457}  &  \makeDataBar{0.332}  &  \makeDataBar{0.357}  \\
{\em rel2-b}  &  \makeDataBar{0.256}  &  \makeDataBar{0.910}  &  \makeDataBar{0.132}  &  \makeDataBar{0.339}  &  \makeDataBar{0.002}  &  \makeDataBar{0.529}  &  \makeDataBar{0.116}  &  \makeDataBar{0.004}  &  \makeDataBar{0.049}  &  \makeDataBar{0.137}  &  \makeDataBar{0.187}  &  \makeDataBar{0.342}  \\
{\em rel2-eb}  &  {\bf \makeDataBar{1.000}}  &  \makeDataBar{0.875}  &  \makeDataBar{0.824}  &  \makeDataBar{0.062}  &  \makeDataBar{0.000}  &  \makeDataBar{0.519}  &  \makeDataBar{0.124}  &  \makeDataBar{0.018}  &  \makeDataBar{0.233}  &  \makeDataBar{0.479}  &  \makeDataBar{0.205}  &  \makeDataBar{0.333}  \\ \hline
{\em abs-c}  &  \makeDataBar{0.021}  &  \makeDataBar{0.037}  &  \makeDataBar{0.000}  &  \makeDataBar{0.000}  &  \makeDataBar{0.000}  &  \makeDataBar{0.506}  &  \makeDataBar{0.000}  &  \makeDataBar{0.005}  &  \makeDataBar{0.250}  &  \makeDataBar{0.420}  &  \makeDataBar{0.550}  &  \makeDataBar{0.312}  \\
{\em rel-eb-c}  &  \makeDataBar{0.006}  &  \makeDataBar{0.027}  &  \makeDataBar{0.504}  &  \makeDataBar{0.721}  &  \makeDataBar{0.000}  &  {\bf \makeDataBar{1.000}}  &  \makeDataBar{0.031}  &  \makeDataBar{0.021}  &  \makeDataBar{0.361}  &  \makeDataBar{0.562}  &  {\bf \makeDataBar{0.581}}  &  \makeDataBar{0.351}   \\
{\em rel2-eb-c}  &  \makeDataBar{0.998}  &  \makeDataBar{0.842}  &  \makeDataBar{0.861}  &  \makeDataBar{0.683}  &  \makeDataBar{0.000}  &  {\bf \makeDataBar{1.000}}  &  \makeDataBar{0.082}  &  \makeDataBar{0.014}  &  \makeDataBar{0.346}  &  \makeDataBar{0.581}  &  \makeDataBar{0.576}  &  {\bf \makeDataBar{0.369}}   \\ \hline
{\em small-2}  &  \makeDataBar{0.998}  &  \makeDataBar{0.842}  &  \makeDataBar{0.861}  &  \makeDataBar{0.683}  &  \makeDataBar{0.000}  &  {\bf \makeDataBar{1.000}}  &  \makeDataBar{0.082}  &  \makeDataBar{0.014}  &  \makeDataBar{0.346}  &  \makeDataBar{0.581}  &  \makeDataBar{0.576}  &  {\bf \makeDataBar{0.369}}  \\
{\em small-4}  &  \makeDataBar{0.992}  &  \makeDataBar{0.877}  &  \makeDataBar{0.939}  &  \makeDataBar{0.805}  &  \makeDataBar{0.000}  &  \makeDataBar{0.500}  &  \makeDataBar{0.197}  &  \makeDataBar{0.001}  &  \makeDataBar{0.509}  &  \makeDataBar{0.734}  &  \makeDataBar{0.520}  &  \makeDataBar{0.342}  \\
{\em small-6}  &  {\bf \makeDataBar{1.000}}  &  \makeDataBar{0.922}  &  \makeDataBar{0.576}  &  \makeDataBar{0.000}  &  \makeDataBar{0.000}  &  \makeDataBar{0.500}  &  {\bf \makeDataBar{0.199}}  &  \makeDataBar{0.007}  &  \makeDataBar{0.571}  &  \makeDataBar{0.766}  &  \makeDataBar{0.516}  &  \makeDataBar{0.330}  \\ \hline
{\em large-2}  &  \makeDataBar{0.998}  &  \makeDataBar{0.896}  &  \makeDataBar{0.933}  &  \makeDataBar{0.882}  &  \makeDataBar{0.000}  &  \makeDataBar{0.500}  &  \makeDataBar{0.197}  &  \makeDataBar{0.002}  &  \makeDataBar{0.548}  &  \makeDataBar{0.762}  &  \makeDataBar{0.530}  &  \makeDataBar{0.314}  \\
{\em large-4}  &  \makeDataBar{0.996}  &  \makeDataBar{0.953}  &  \makeDataBar{0.848}  &  \makeDataBar{0.855}  &  \makeDataBar{0.000}  &  \makeDataBar{0.500}  &  {\bf \makeDataBar{0.199}}  &  \makeDataBar{0.010}  &  \makeDataBar{0.523}  &  \makeDataBar{0.782}  &  \makeDataBar{0.500}  &  \makeDataBar{0.360}  \\
{\em large-6}  &  \makeDataBar{0.994}  &  \makeDataBar{0.887}  &  \makeDataBar{0.619}  &  \makeDataBar{0.856}  &  \makeDataBar{0.000}  &  \makeDataBar{0.500}  &  \makeDataBar{0.195}  &  \makeDataBar{0.000}  &  \makeDataBar{0.549}  &  \makeDataBar{0.766}  &  \makeDataBar{0.483}  &  \makeDataBar{0.317}  \\ \hline
{\em small-2s}  &  \makeDataBar{0.998}  &  \makeDataBar{0.871}  &  {\bf \makeDataBar{0.979}}  &  {\bf \makeDataBar{0.972}}  &  \makeDataBar{0.000}  &  {\bf \makeDataBar{1.000}}  &  \makeDataBar{0.044}  &  \makeDataBar{0.006}  &  \makeDataBar{0.328}  &  \makeDataBar{0.519}  &  \makeDataBar{0.487}  &  \makeDataBar{0.348}  \\
{\em small-4s}  &  \makeDataBar{0.998}  &  \makeDataBar{0.986}  &  \makeDataBar{0.870}  &  \makeDataBar{0.871}  &  \makeDataBar{0.000}  &  \makeDataBar{0.500}  &  \makeDataBar{0.175}  &  {\bf \makeDataBar{0.039}}  &  \makeDataBar{0.540}  &  \makeDataBar{0.742}  &  \makeDataBar{0.515}  &  \makeDataBar{0.362}  \\
{\em small-6s}  &  {\bf \makeDataBar{1.000}}  &  \makeDataBar{0.984}  &  \makeDataBar{0.821}  &  \makeDataBar{0.000}  &  \makeDataBar{0.000}  &  \makeDataBar{0.500}  &  {\bf \makeDataBar{0.199}}  &  \makeDataBar{0.000}  &  \makeDataBar{0.569}  &  \makeDataBar{0.788}  &  \makeDataBar{0.486}  &  \makeDataBar{0.344}  \\ \hline
{\em large-2s}  &  {\bf \makeDataBar{1.000}}  &  \makeDataBar{0.945}  &  \makeDataBar{0.952}  &  \makeDataBar{0.955}  &  \makeDataBar{0.000}  &  {\bf \makeDataBar{1.000}}  &  \makeDataBar{0.054}  &  \makeDataBar{0.003}  &  \makeDataBar{0.526}  &  \makeDataBar{0.641}  &  \makeDataBar{0.563}  &  \makeDataBar{0.304} \\
{\em large-4s}  &  {\bf \makeDataBar{1.000}}  &  \makeDataBar{0.959}  &  \makeDataBar{0.923}  &  \makeDataBar{0.959}  &  \makeDataBar{0.000}  &  \makeDataBar{0.500}  &  \makeDataBar{0.195}  &  \makeDataBar{0.004}  &  \makeDataBar{0.604}  &  \makeDataBar{0.810}  &  \makeDataBar{0.481}  &  \makeDataBar{0.362}  \\
{\em large-6s}  &  {\bf \makeDataBar{1.000}}  &  {\bf \makeDataBar{0.998}}  &  \makeDataBar{0.489}  &  \makeDataBar{0.000}  &  \makeDataBar{0.000}  &  \makeDataBar{0.500}  &  \makeDataBar{0.198}  &  \makeDataBar{0.000}  &  {\bf \makeDataBar{0.642}}  &  {\bf \makeDataBar{0.832}}  &  \makeDataBar{0.469}  &  \makeDataBar{0.361}  \\
\end{tabular}		
}
\caption{Maximum sequence-level accuracy achieved in a given repetition for all the models evaluated in this paper.}
\label{tbl:all-max} 
\end{table*}

\begin{table*}[tb]\centering 
\resizebox{\textwidth}{!}{
\setlength{\tabcolsep}{4pt}
\begin{tabular}{l|cccccc|cccccc} 
& Add & AddNeg & Reverse & Dup & Cart & Inters & SCAN-l & SCAN-aj & PCFG-p & PCFG-s & COGS & CFQ \\ \hline
{\em abs}  &  \makeDataBar{0.003}  &  \makeDataBar{0.047}  &  \makeDataBar{0.001}  &  \makeDataBar{0.000}  &  \makeDataBar{0.000}  &  \makeDataBar{0.000}  &  \makeDataBar{0.000}  &  \makeDataBar{0.004}  &  \makeDataBar{0.014}  &  \makeDataBar{0.039}  &  \makeDataBar{0.067}  &  \makeDataBar{0.022} \\ \hline
{\em rel-e}  &  \makeDataBar{0.003}  &  \makeDataBar{0.017}  &  \makeDataBar{0.169}  &  \makeDataBar{0.271}  &  \makeDataBar{0.012}  &  \makeDataBar{0.003}  &  \makeDataBar{0.045}  &  \makeDataBar{0.004}  &  \makeDataBar{0.023}  &  \makeDataBar{0.078}  &  \makeDataBar{0.103}  &  \makeDataBar{0.027} \\
{\em rel-b}  &  \makeDataBar{0.002}  &  \makeDataBar{0.006}  &  \makeDataBar{0.078}  &  \makeDataBar{0.046}  &  \makeDataBar{0.073}  &  \makeDataBar{0.003}  &  \makeDataBar{0.038}  &  \makeDataBar{0.006}  &  \makeDataBar{0.005}  &  \makeDataBar{0.014}  &  \makeDataBar{0.040}  &  \makeDataBar{0.025} \\
{\em rel-eb}  &  \makeDataBar{0.002}  &  \makeDataBar{0.007}  &  \makeDataBar{0.211}  &  \makeDataBar{0.287}  &  \makeDataBar{0.000}  &  \makeDataBar{0.001}  &  \makeDataBar{0.038}  &  \makeDataBar{0.011}  &  \makeDataBar{0.013}  &  \makeDataBar{0.066}  &  \makeDataBar{0.050}  &  \makeDataBar{0.047} \\ \hline
{\em rel2-e}  &  \makeDataBar{0.009}  &  \makeDataBar{0.074}  &  \makeDataBar{0.167}  &  \makeDataBar{0.014}  &  \makeDataBar{0.000}  &  \makeDataBar{0.004}  &  \makeDataBar{0.023}  &  \makeDataBar{0.009}  &  \makeDataBar{0.016}  &  \makeDataBar{0.111}  &  \makeDataBar{0.104}  &  \makeDataBar{0.035} \\
{\em rel2-b}  &  \makeDataBar{0.122}  &  \makeDataBar{0.202}  &  \makeDataBar{0.051}  &  \makeDataBar{0.055}  &  \makeDataBar{0.001}  &  \makeDataBar{0.009}  &  \makeDataBar{0.039}  &  \makeDataBar{0.002}  &  \makeDataBar{0.011}  &  \makeDataBar{0.018}  &  \makeDataBar{0.045}  &  \makeDataBar{0.016} \\
{\em rel2-eb}  &  \makeDataBar{0.029}  &  \makeDataBar{0.067}  &  \makeDataBar{0.057}  &  \makeDataBar{0.024}  &  \makeDataBar{0.000}  &  \makeDataBar{0.007}  &  \makeDataBar{0.029}  &  \makeDataBar{0.008}  &  \makeDataBar{0.047}  &  \makeDataBar{0.101}  &  \makeDataBar{0.043}  &  \makeDataBar{0.020} \\ \hline
{\em abs-c}  &  \makeDataBar{0.009}  &  \makeDataBar{0.010}  &  \makeDataBar{0.000}  &  \makeDataBar{0.000}  &  \makeDataBar{0.000}  &  \makeDataBar{0.003}  &  \makeDataBar{0.000}  &  \makeDataBar{0.002}  &  \makeDataBar{0.024}  &  \makeDataBar{0.027}  &  \makeDataBar{0.038}  &  \makeDataBar{0.013} \\
{\em rel-eb-c}  &  \makeDataBar{0.003}  &  \makeDataBar{0.011}  &  \makeDataBar{0.135}  &  \makeDataBar{0.157}  &  \makeDataBar{0.000}  &  \makeDataBar{0.322}  &  \makeDataBar{0.005}  &  \makeDataBar{0.011}  &  \makeDataBar{0.017}  &  \makeDataBar{0.036}  &  \makeDataBar{0.093}  &  \makeDataBar{0.025} \\
{\em rel2-eb-c}  &  \makeDataBar{0.035}  &  \makeDataBar{0.053}  &  \makeDataBar{0.208}  &  \makeDataBar{0.289}  &  \makeDataBar{0.000}  &  \makeDataBar{0.239}  &  \makeDataBar{0.033}  &  \makeDataBar{0.005}  &  \makeDataBar{0.009}  &  \makeDataBar{0.048}  &  \makeDataBar{0.056}  &  \makeDataBar{0.063} \\ \hline
{\em small-2}  &  \makeDataBar{0.035}  &  \makeDataBar{0.053}  &  \makeDataBar{0.208}  &  \makeDataBar{0.289}  &  \makeDataBar{0.000}  &  \makeDataBar{0.239}  &  \makeDataBar{0.033}  &  \makeDataBar{0.005}  &  \makeDataBar{0.009}  &  \makeDataBar{0.048}  &  \makeDataBar{0.056}  &  \makeDataBar{0.063} \\ 
{\em small-4}  &  \makeDataBar{0.004}  &  \makeDataBar{0.054}  &  \makeDataBar{0.213}  &  \makeDataBar{0.184}  &  \makeDataBar{0.000}  &  \makeDataBar{0.000}  &  \makeDataBar{0.046}  &  \makeDataBar{0.000}  &  \makeDataBar{0.010}  &  \makeDataBar{0.019}  &  \makeDataBar{0.028}  &  \makeDataBar{0.049} \\
{\em small-6}  &  \makeDataBar{0.007}  &  \makeDataBar{0.120}  &  \makeDataBar{0.233}  &  \makeDataBar{0.000}  &  \makeDataBar{0.000}  &  \makeDataBar{0.256}  &  \makeDataBar{0.056}  &  \makeDataBar{0.004}  &  \makeDataBar{0.024}  &  \makeDataBar{0.026}  &  \makeDataBar{0.047}  &  \makeDataBar{0.022} \\ \hline
{\em large-2}  &  \makeDataBar{0.016}  &  \makeDataBar{0.074}  &  \makeDataBar{0.240}  &  \makeDataBar{0.289}  &  \makeDataBar{0.000}  &  \makeDataBar{0.000}  &  \makeDataBar{0.022}  &  \makeDataBar{0.001}  &  \makeDataBar{0.012}  &  \makeDataBar{0.004}  &  \makeDataBar{0.042}  &  \makeDataBar{0.033} \\
{\em large-4}  &  \makeDataBar{0.075}  &  \makeDataBar{0.106}  &  \makeDataBar{0.178}  &  \makeDataBar{0.190}  &  \makeDataBar{0.000}  &  \makeDataBar{0.211}  &  \makeDataBar{0.049}  &  \makeDataBar{0.006}  &  \makeDataBar{0.009}  &  \makeDataBar{0.010}  &  \makeDataBar{0.033}  &  \makeDataBar{0.047} \\
{\em large-6}  &  \makeDataBar{0.023}  &  \makeDataBar{0.377}  &  \makeDataBar{0.119}  &  \makeDataBar{0.356}  &  \makeDataBar{0.000}  &  \makeDataBar{0.264}  &  \makeDataBar{0.045}  &  \makeDataBar{0.000}  &  \makeDataBar{0.018}  &  \makeDataBar{0.014}  &  \makeDataBar{0.029}  &  \makeDataBar{0.022} \\ \hline
{\em small-2s}  &  \makeDataBar{0.007}  &  \makeDataBar{0.038}  &  \makeDataBar{0.255}  &  \makeDataBar{0.254}  &  \makeDataBar{0.000}  &  \makeDataBar{0.346}  &  \makeDataBar{0.021}  &  \makeDataBar{0.003}  &  \makeDataBar{0.014}  &  \makeDataBar{0.019}  &  \makeDataBar{0.054}  &  \makeDataBar{0.039} \\
{\em small-4s}  &  \makeDataBar{0.009}  &  \makeDataBar{0.055}  &  \makeDataBar{0.118}  &  \makeDataBar{0.261}  &  \makeDataBar{0.000}  &  \makeDataBar{0.000}  &  \makeDataBar{0.005}  &  \makeDataBar{0.020}  &  \makeDataBar{0.008}  &  \makeDataBar{0.023}  &  \makeDataBar{0.068}  &  \makeDataBar{0.054} \\
{\em small-6s}  &  \makeDataBar{0.012}  &  \makeDataBar{0.047}  &  \makeDataBar{0.208}  &  \makeDataBar{0.000}  &  \makeDataBar{0.000}  &  \makeDataBar{0.001}  &  \makeDataBar{0.017}  &  \makeDataBar{0.000}  &  \makeDataBar{0.006}  &  \makeDataBar{0.007}  &  \makeDataBar{0.030}  &  \makeDataBar{0.041} \\ \hline
{\em large-2s}  &  \makeDataBar{0.004}  &  \makeDataBar{0.031}  &  \makeDataBar{0.131}  &  \makeDataBar{0.167}  &  \makeDataBar{0.000}  &  \makeDataBar{0.156}  &  \makeDataBar{0.027}  &  \makeDataBar{0.001}  &  \makeDataBar{0.018}  &  \makeDataBar{0.004}  &  \makeDataBar{0.102}  &  \makeDataBar{0.011} \\
{\em large-4s}  &  \makeDataBar{0.007}  &  \makeDataBar{0.039}  &  \makeDataBar{0.127}  &  \makeDataBar{0.066}  &  \makeDataBar{0.000}  &  \makeDataBar{0.211}  &  \makeDataBar{0.016}  &  \makeDataBar{0.002}  &  \makeDataBar{0.015}  &  \makeDataBar{0.017}  &  \makeDataBar{0.009}  &  \makeDataBar{0.043} \\
{\em large-6s}  &  \makeDataBar{0.020}  &  \makeDataBar{0.015}  &  \makeDataBar{0.159}  &  \makeDataBar{0.000}  &  \makeDataBar{0.000}  &  \makeDataBar{0.000}  &  \makeDataBar{0.002}  &  \makeDataBar{0.000}  &  \makeDataBar{0.008}  &  \makeDataBar{0.007}  &  \makeDataBar{0.013}  &  \makeDataBar{0.037} \\
\end{tabular}		
}
\caption{Standard deviation of the sequence level accuracy results.}
\label{tbl:all-stddev} 
\end{table*}

\section{Parameter Counts}

Table \ref{tbl:parameters} shows the parameter count for all the models used in this paper, notice that exact parameter counts vary per dataset, as each dataset has a different token vocabulary, and hence both the token embedding and the output layers vary. One interesting result is that in our experiments, parameter count is not, by itself, sufficient to increase compositional generalization. Our best model overall ({\em large-4s}) only had about 0.5 million parameters, and outperformed significantly larger models. Another example, of this is that the models with shared layer parameters outperform their counterparts without parameter sharing, although they naturally have less parameters.

\begin{table*}[tb]\centering 
\resizebox{\textwidth}{!}{
\setlength{\tabcolsep}{4pt}
\begin{tabular}{l|cccccc|cccccc} 
& Add & AddNeg & Reverse & Dup & Cart & Inters & SCAN-l & SCAN-aj & PCFG-p & PCFG-s & COGS & CFQ \\ \hline
{\em abs}  &  236k  &  236k  &  236k  &  236k  &  238k  &  253k  &  238k  &  238k  &  337k  &  337k  &  402k  &  268k \\ \hline
{\em rel-e}  &  239k  &  239k  &  239k  &  239k  &  241k  &  257k  &  241k  &  241k  &  340k  &  340k  &  405k  &  272k \\
{\em rel-b}  &  236k  &  236k  &  236k  &  236k  &  238k  &  254k  &  238k  &  238k  &  337k  &  337k  &  402k  &  269k \\
{\em rel-eb}  &  239k  &  239k  &  239k  &  239k  &  241k  &  257k  &  241k  &  241k  &  340k  &  340k  &  405k  &  272k \\ \hline
{\em rel2-e}  &  239k  &  239k  &  239k  &  239k  &  241k  &  257k  &  241k  &  241k  &  340k  &  340k  &  405k  &  272k \\
{\em rel2-b}  &  236k  &  236k  &  236k  &  236k  &  238k  &  254k  &  238k  &  238k  &  337k  &  337k  &  402k  &  269k \\
{\em rel2-eb}  &  239k  &  239k  &  239k  &  239k  &  241k  &  257k  &  241k  &  241k  &  340k  &  340k  &  405k  &  272k \\ \hline
{\em abs-c}  &  241k  &  241k  &  241k  &  241k  &  242k  &  258k  &  243k  &  243k  &  341k  &  341k  &  407k  &  273k \\
{\em rel-eb-c}  &  243k  &  244k  &  243k  &  243k  &  245k  &  261k  &  245k  &  245k  &  344k  &  344k  &  410k  &  276k \\
{\em rel2-eb-c}  &  243k  &  244k  &  243k  &  243k  &  245k  &  261k  &  245k  &  245k  &  344k  &  344k  &  410k  &  276k \\ \hline
{\em small-2}  &  243k  &  244k  &  243k  &  243k  &  245k  &  261k  &  245k  &  245k  &  344k  &  344k  &  410k  &  276k \\ 
{\em small-4}  &  480k  &  480k  &  480k  &  480k  &  482k  &  498k  &  482k  &  482k  &  581k  &  581k  &  646k  &  513k \\
{\em small-6}  &  717k  &  717k  &  717k  &  717k  &  719k  &  735k  &  719k  &  719k  &  818k  &  818k  &  883k  &  750k \\ \hline
{\em large-2}  &  1.88m  &  1.88m  &  1.88m  &  1.88m  &  1.88m  &  1.92m  &  1.88m  &  1.88m  &  2.08m  &  2.08m  &  2.21m  &  1.95m \\
{\em large-4}  &  1.88m  &  1.88m  &  1.88m  &  1.88m  &  1.88m  &  1.92m  &  1.88m  &  1.88m  &  2.08m  &  2.08m  &  2.21m  &  1.95m \\
{\em large-6}  &  2.81m  &  2.81m  &  2.81m  &  2.81m  &  2.81m  &  2.84m  &  2.81m  &  2.81m  &  3.01m  &  3.01m  &  3.14m  &  2.87m \\ \hline
{\em small-2s}  &  125k  &  125k  &  125k  &  125k  &  127k  &  143k  &  127k  &  127k  &  226k  &  226k  &  291k  &  158k \\
{\em small-4s}  &  125k  &  125k  &  125k  &  125k  &  127k  &  143k  &  127k  &  127k  &  226k  &  226k  &  291k  &  158k \\
{\em small-6s}  &  125k  &  125k  &  125k  &  125k  &  127k  &  143k  &  127k  &  127k  &  226k  &  226k  &  291k  &  158k \\ \hline
{\em large-2s}  &  486k  &  487k  &  486k  &  486k  &  490k  &  521k  &  490k  &  490k  &  687k  &  687k  &  818k  &  552k \\
{\em large-4s}  &  486k  &  487k  &  486k  &  486k  &  490k  &  521k  &  490k  &  490k  &  687k  &  687k  &  818k  &  552k \\
{\em large-6s}  &  486k  &  487k  &  486k  &  486k  &  490k  &  521k  &  490k  &  490k  &  687k  &  687k  &  818k  &  552k \\
\end{tabular}		
}
\caption{Parameter counts for the models used in this paper.}
\label{tbl:parameters} 
\end{table*}

\section{Detailed Results in COGS}

Table \ref{tbl:cogs-full} shows the results of some of the models we tested in the COGS dataset (including seq2seq and sequence tagging models), with the accuracy broken down by the type of example in the generalization set. The COGS dataset contains four splits: training, dev, test and generalization (generalization is the one used to measure compositional generalization, and the set reported in the main paper). All but one shown configuration achieve more than 95\% sequence level accuracy on the test and development splits after training for 16 epochs over the training data. The generalization set is split into several generalization tasks as described above, to break down performance by type of  generalization (overall performance in the generalization set is shown in the bottom row).

The best tagging model does much better than the base seq2seq model (0.784 vs. 0.278). Notably the tagging model does relatively well on the {\em Depth generalization: Prepositional phrase (PP) modifiers} task achieving accuracy 0.681. When the depth of the model is increased from 2 to 6, the score on this task increases from 0.681 to 0.819, i.e. the model with more layers can parse deeper recursion. However, increasing the encoder depth at the same time dramatically lowers the performance on {\em Verb Argument Structure Alternation} tasks.

Since many of the tasks are solved to near perfect accuracy, here we briefly discuss the types of the remaining errors. The one type of task where sequence tagging models did worse than seq2seq is {\em Prim verb $\to$ Infinitival argument}, which measures one shot generalization of an example with only a single verb to examples where the verb is used in sentences. The cause of this is that the tagging example with only a single verb doesn't actually encode the type of relations the verb allows, so the tagging model is actually not provided the full information in the only example for this one shot learning task. 
Nevertheless, this category was solved in our seq2seq models with a copy decoder.

Curiously, some errors, that the tagging model with {\em attention} in the parent prediction head makes, are quite quite reasonable. For example in the {\em Obj-mod PP $\to$ Subj-mod PP} task, the model often gets the complete parsing tree correctly, and the only error is the predicted relation of the subject to the predicate (instead of {\em agent} the model predicts {\em theme} as is present in all the training examples, where the prepositional phrase modifies the object). 

Another task where even the best tagging model achieves a low score (0.233) is {\em Depth generalization: Sentential complements}. The examples in this task are long complex sentences chained together with the conjunction {\em that}. The most common error here is to predict that the main verb depends on another verb far away in the sentence structure, instead of predicting that it has no parent. The distance to the incorrectly predicted parent is often more than 16, which was the limit on our relative attention offsets. The attention mechanism seems to get confused by seeing many more tokens in this test split than during training.

\begin{sidewaystable*}[tb]\centering 
\resizebox{\textwidth}{!}{
\setlength{\tabcolsep}{4pt}
\begin{tabular}{l|cccccc|cccccccccccc}
& \multicolumn{6}{c|}{seq2seq} & \multicolumn{12}{c}{tagging} \\ \hline
Model & abs & abs & rel2-eb & rel2-eb & rel2-eb-c & rel2-eb-c & abs & abs & abs & abs & abs & abs & rel-eb & rel-eb & rel-eb & rel-eb & rel-eb & rel-eb \\
Size & small-2 & small-6 & small-2s & small-6s & small-2s & small-6s & small-2 & small-6 & small-2 & small-6 & small-2 & small-6 & small-2s & small-6s & small-2s & small-6s & small-2s & small-6s \\
Parent encoding &  &  &  &  &  &  & absolute & absolute & relative & relative & attention & attention & absolute & absolute & relative & relative & attention & attention \\ \hline 
Test split & \makeDataBar{0.981} & \makeDataBar{0.646} & \makeDataBar{0.978} & \makeDataBar{0.961} & \makeDataBar{0.983} & \makeDataBar{0.974} & \makeDataBar{0.997} & \makeDataBar{0.994} & \makeDataBar{0.997} & \makeDataBar{0.997} & \makeDataBar{0.997} & \makeDataBar{0.997} & \makeDataBar{0.995} & \makeDataBar{0.997} & \makeDataBar{1.000} & \makeDataBar{1.000} & \makeDataBar{1.000} & \makeDataBar{1.000} \\
Dev split & \makeDataBar{0.976} & \makeDataBar{0.625} & \makeDataBar{0.968} & \makeDataBar{0.950} & \makeDataBar{0.976} & \makeDataBar{0.974} & \makeDataBar{0.996} & \makeDataBar{0.994} & \makeDataBar{0.997} & \makeDataBar{0.996} & \makeDataBar{0.997} & \makeDataBar{0.996} & \makeDataBar{0.995} & \makeDataBar{0.996} & \makeDataBar{1.000} & \makeDataBar{1.000} & \makeDataBar{1.000} & \makeDataBar{0.999} \\
\hline \hline 
\multicolumn{17}{l}{ Lexical Generalization: Novel Combination of Familiar Primitives and Grammatical Roles } \\ \hline 
Subject $\to$ Object (common noun) & \makeDataBar{0.309} & \makeDataBar{0.008} & \makeDataBar{0.030} & \makeDataBar{0.011} & \makeDataBar{0.900} & \makeDataBar{0.899} & \makeDataBar{0.911} & \makeDataBar{0.938} & \makeDataBar{0.914} & \makeDataBar{0.916} & \makeDataBar{0.893} & \makeDataBar{0.918} & \makeDataBar{0.899} & \makeDataBar{0.972} & \makeDataBar{0.978} & \makeDataBar{0.996} & \makeDataBar{0.969} & \makeDataBar{0.956} \\
Subject $\to$ Object (proper noun) & \makeDataBar{0.098} & \makeDataBar{0.000} & \makeDataBar{0.000} & \makeDataBar{0.000} & \makeDataBar{0.581} & \makeDataBar{0.429} & \makeDataBar{0.630} & \makeDataBar{0.590} & \makeDataBar{0.563} & \makeDataBar{0.494} & \makeDataBar{0.731} & \makeDataBar{0.610} & \makeDataBar{0.690} & \makeDataBar{0.671} & \makeDataBar{0.567} & \makeDataBar{0.580} & \makeDataBar{0.826} & \makeDataBar{0.672} \\
Object $\to$ Subject (common noun) & \makeDataBar{0.790} & \makeDataBar{0.091} & \makeDataBar{0.304} & \makeDataBar{0.175} & \makeDataBar{0.959} & \makeDataBar{0.936} & \makeDataBar{0.982} & \makeDataBar{0.973} & \makeDataBar{0.974} & \makeDataBar{0.994} & \makeDataBar{0.965} & \makeDataBar{0.935} & \makeDataBar{0.945} & \makeDataBar{0.988} & \makeDataBar{0.992} & \makeDataBar{0.999} & \makeDataBar{0.978} & \makeDataBar{0.769} \\
Object $\to$ Subject (proper noun) & \makeDataBar{0.207} & \makeDataBar{0.007} & \makeDataBar{0.023} & \makeDataBar{0.019} & \makeDataBar{0.970} & \makeDataBar{0.951} & \makeDataBar{0.993} & \makeDataBar{0.986} & \makeDataBar{0.995} & \makeDataBar{0.991} & \makeDataBar{0.993} & \makeDataBar{0.990} & \makeDataBar{0.985} & \makeDataBar{0.847} & \makeDataBar{0.998} & \makeDataBar{0.999} & \makeDataBar{0.995} & \makeDataBar{0.984} \\
Primitive noun $\to$ Subject (common noun) & \makeDataBar{0.240} & \makeDataBar{0.098} & \makeDataBar{0.242} & \makeDataBar{0.216} & \makeDataBar{0.956} & \makeDataBar{0.913} & \makeDataBar{0.993} & \makeDataBar{0.983} & \makeDataBar{0.995} & \makeDataBar{0.972} & \makeDataBar{0.990} & \makeDataBar{0.978} & \makeDataBar{0.927} & \makeDataBar{0.976} & \makeDataBar{1.000} & \makeDataBar{1.000} & \makeDataBar{0.988} & \makeDataBar{0.927} \\
Primitive noun $\to$ Subject (proper noun) & \makeDataBar{0.019} & \makeDataBar{0.004} & \makeDataBar{0.016} & \makeDataBar{0.017} & \makeDataBar{0.422} & \makeDataBar{0.772} & \makeDataBar{0.974} & \makeDataBar{0.983} & \makeDataBar{0.979} & \makeDataBar{0.985} & \makeDataBar{0.993} & \makeDataBar{0.992} & \makeDataBar{0.796} & \makeDataBar{0.990} & \makeDataBar{0.959} & \makeDataBar{0.803} & \makeDataBar{0.996} & \makeDataBar{0.999} \\
Primitive noun $\to$ Object (common noun) & \makeDataBar{0.017} & \makeDataBar{0.007} & \makeDataBar{0.014} & \makeDataBar{0.012} & \makeDataBar{0.929} & \makeDataBar{0.902} & \makeDataBar{0.950} & \makeDataBar{0.968} & \makeDataBar{0.939} & \makeDataBar{0.945} & \makeDataBar{0.949} & \makeDataBar{0.972} & \makeDataBar{0.904} & \makeDataBar{0.986} & \makeDataBar{0.996} & \makeDataBar{1.000} & \makeDataBar{0.953} & \makeDataBar{0.966} \\
Primitive noun $\to$ Object (proper noun) & \makeDataBar{0.000} & \makeDataBar{0.000} & \makeDataBar{0.000} & \makeDataBar{0.000} & \makeDataBar{0.200} & \makeDataBar{0.513} & \makeDataBar{0.651} & \makeDataBar{0.687} & \makeDataBar{0.557} & \makeDataBar{0.569} & \makeDataBar{0.722} & \makeDataBar{0.649} & \makeDataBar{0.624} & \makeDataBar{0.676} & \makeDataBar{0.545} & \makeDataBar{0.467} & \makeDataBar{0.700} & \makeDataBar{0.673} \\
Primitive verb $\to$ Infinitival argument & \makeDataBar{0.000} & \makeDataBar{0.000} & \makeDataBar{0.000} & \makeDataBar{0.000} & \makeDataBar{0.476} & \makeDataBar{0.766} & \makeDataBar{0.000} & \makeDataBar{0.000} & \makeDataBar{0.000} & \makeDataBar{0.000} & \makeDataBar{0.011} & \makeDataBar{0.000} & \makeDataBar{0.017} & \makeDataBar{0.000} & \makeDataBar{0.000} & \makeDataBar{0.000} & \makeDataBar{0.001} & \makeDataBar{0.000} \\
\hline \hline 
\multicolumn{17}{l}{ Lexical Generalization: Verb Argument Structure Alternation } \\ \hline 
Active $\to$ Passive & \makeDataBar{0.604} & \makeDataBar{0.107} & \makeDataBar{0.147} & \makeDataBar{0.000} & \makeDataBar{0.000} & \makeDataBar{0.000} & \makeDataBar{0.697} & \makeDataBar{0.122} & \makeDataBar{0.741} & \makeDataBar{0.160} & \makeDataBar{0.736} & \makeDataBar{0.004} & \makeDataBar{0.210} & \makeDataBar{0.013} & \makeDataBar{0.612} & \makeDataBar{0.001} & \makeDataBar{0.948} & \makeDataBar{0.000} \\
Passive $\to$ Active & \makeDataBar{0.196} & \makeDataBar{0.067} & \makeDataBar{0.002} & \makeDataBar{0.001} & \makeDataBar{0.001} & \makeDataBar{0.001} & \makeDataBar{0.535} & \makeDataBar{0.115} & \makeDataBar{0.617} & \makeDataBar{0.014} & \makeDataBar{0.625} & \makeDataBar{0.163} & \makeDataBar{0.539} & \makeDataBar{0.000} & \makeDataBar{0.586} & \makeDataBar{0.073} & \makeDataBar{0.897} & \makeDataBar{0.000} \\
Object-omitted transitive $\to$ Transitive & \makeDataBar{0.275} & \makeDataBar{0.002} & \makeDataBar{0.002} & \makeDataBar{0.001} & \makeDataBar{0.001} & \makeDataBar{0.003} & \makeDataBar{0.527} & \makeDataBar{0.100} & \makeDataBar{0.620} & \makeDataBar{0.031} & \makeDataBar{0.413} & \makeDataBar{0.092} & \makeDataBar{0.356} & \makeDataBar{0.000} & \makeDataBar{0.447} & \makeDataBar{0.027} & \makeDataBar{0.926} & \makeDataBar{0.000} \\
Unaccusative $\to$ Transitive & \makeDataBar{0.069} & \makeDataBar{0.000} & \makeDataBar{0.002} & \makeDataBar{0.001} & \makeDataBar{0.002} & \makeDataBar{0.003} & \makeDataBar{0.528} & \makeDataBar{0.144} & \makeDataBar{0.295} & \makeDataBar{0.000} & \makeDataBar{0.457} & \makeDataBar{0.000} & \makeDataBar{0.301} & \makeDataBar{0.000} & \makeDataBar{0.326} & \makeDataBar{0.005} & \makeDataBar{0.787} & \makeDataBar{0.005} \\
Double object dative $\to$ PP dative & \makeDataBar{0.819} & \makeDataBar{0.005} & \makeDataBar{0.095} & \makeDataBar{0.001} & \makeDataBar{0.002} & \makeDataBar{0.000} & \makeDataBar{0.590} & \makeDataBar{0.279} & \makeDataBar{0.778} & \makeDataBar{0.345} & \makeDataBar{0.732} & \makeDataBar{0.037} & \makeDataBar{0.424} & \makeDataBar{0.000} & \makeDataBar{0.853} & \makeDataBar{0.018} & \makeDataBar{0.958} & \makeDataBar{0.000} \\
PP dative $\to$ Double object dative & \makeDataBar{0.404} & \makeDataBar{0.002} & \makeDataBar{0.053} & \makeDataBar{0.003} & \makeDataBar{0.006} & \makeDataBar{0.004} & \makeDataBar{0.771} & \makeDataBar{0.542} & \makeDataBar{0.617} & \makeDataBar{0.169} & \makeDataBar{0.895} & \makeDataBar{0.042} & \makeDataBar{0.616} & \makeDataBar{0.008} & \makeDataBar{0.717} & \makeDataBar{0.069} & \makeDataBar{0.850} & \makeDataBar{0.139} \\
\hline \hline 
\multicolumn{17}{l}{ Lexical Generalization: Verb Class } \\ \hline 
Agent NP $\to$ Unaccusative Subject & \makeDataBar{0.399} & \makeDataBar{0.000} & \makeDataBar{0.003} & \makeDataBar{0.002} & \makeDataBar{0.958} & \makeDataBar{0.951} & \makeDataBar{0.784} & \makeDataBar{0.661} & \makeDataBar{0.951} & \makeDataBar{0.911} & \makeDataBar{0.952} & \makeDataBar{0.949} & \makeDataBar{0.982} & \makeDataBar{0.980} & \makeDataBar{0.995} & \makeDataBar{0.999} & \makeDataBar{1.000} & \makeDataBar{0.999} \\
Theme NP $\to$ Object-omitted transitive Subject & \makeDataBar{0.688} & \makeDataBar{0.000} & \makeDataBar{0.023} & \makeDataBar{0.000} & \makeDataBar{0.818} & \makeDataBar{0.965} & \makeDataBar{0.791} & \makeDataBar{0.644} & \makeDataBar{0.642} & \makeDataBar{0.861} & \makeDataBar{0.473} & \makeDataBar{0.420} & \makeDataBar{0.537} & \makeDataBar{0.831} & \makeDataBar{0.984} & \makeDataBar{0.903} & \makeDataBar{0.701} & \makeDataBar{0.485} \\
Theme NP $\to$ Unergative subject & \makeDataBar{0.694} & \makeDataBar{0.000} & \makeDataBar{0.023} & \makeDataBar{0.000} & \makeDataBar{0.843} & \makeDataBar{0.966} & \makeDataBar{0.930} & \makeDataBar{0.715} & \makeDataBar{0.643} & \makeDataBar{0.858} & \makeDataBar{0.544} & \makeDataBar{0.570} & \makeDataBar{0.583} & \makeDataBar{0.895} & \makeDataBar{0.960} & \makeDataBar{0.919} & \makeDataBar{0.771} & \makeDataBar{0.530} \\
\hline \hline 
\multicolumn{17}{l}{ Structural Generalization: Novel Combination Modified Phrases and Grammatical Roles } \\ \hline 
Object-modifying PP $\to$ Subject-modifying PP & \makeDataBar{0.000} & \makeDataBar{0.000} & \makeDataBar{0.000} & \makeDataBar{0.000} & \makeDataBar{0.000} & \makeDataBar{0.000} & \makeDataBar{0.000} & \makeDataBar{0.000} & \makeDataBar{0.000} & \makeDataBar{0.000} & \makeDataBar{0.007} & \makeDataBar{0.029} & \makeDataBar{0.000} & \makeDataBar{0.000} & \makeDataBar{0.000} & \makeDataBar{0.000} & \makeDataBar{0.299} & \makeDataBar{0.371} \\
\hline \hline 
\multicolumn{17}{l}{ Structural Generalization: Deeper Recursion } \\ \hline 
Depth generalization: PP modifiers & \makeDataBar{0.003} & \makeDataBar{0.000} & \makeDataBar{0.000} & \makeDataBar{0.000} & \makeDataBar{0.000} & \makeDataBar{0.000} & \makeDataBar{0.138} & \makeDataBar{0.074} & \makeDataBar{0.231} & \makeDataBar{0.191} & \makeDataBar{0.133} & \makeDataBar{0.000} & \makeDataBar{0.000} & \makeDataBar{0.010} & \makeDataBar{0.669} & \makeDataBar{0.775} & \makeDataBar{0.681} & \makeDataBar{0.819} \\
Depth generalization: Sentential complements & \makeDataBar{0.000} & \makeDataBar{0.000} & \makeDataBar{0.000} & \makeDataBar{0.000} & \makeDataBar{0.000} & \makeDataBar{0.000} & \makeDataBar{0.000} & \makeDataBar{0.000} & \makeDataBar{0.017} & \makeDataBar{0.005} & \makeDataBar{0.000} & \makeDataBar{0.000} & \makeDataBar{0.000} & \makeDataBar{0.000} & \makeDataBar{0.282} & \makeDataBar{0.164} & \makeDataBar{0.233} & \makeDataBar{0.133} \\ \hline \hline
Overall & \makeDataBar{0.278} & \makeDataBar{0.019} & \makeDataBar{0.047} & \makeDataBar{0.022} & \makeDataBar{0.430} & \makeDataBar{0.475} & \makeDataBar{0.637} & \makeDataBar{0.500} & \makeDataBar{0.622} & \makeDataBar{0.496} & \makeDataBar{0.629} & \makeDataBar{0.445} & \makeDataBar{0.540} & \makeDataBar{0.469} & \makeDataBar{0.689} & \makeDataBar{0.514} & \makeDataBar{0.784} & \makeDataBar{0.497} \\

\end{tabular}		
}
\caption{Sequence-level accuracy in different subsets of the generalization set in COGS for both seq2seq and sequence tagging models (averaged over 5 runs). PP stands for prepositional phrase.}
\label{tbl:cogs-full} 
\end{sidewaystable*}

\section{Dataset Details}

This appendix presents more details on the datasets used in this paper, as well as on the type of compositionality involved in each of them. 

\begin{itemize}
    \item 
    {\bf Addition} ({\em Add}): This is a synthetic addition task, where the input contains the digits of two integers, and the output should be the digits of their sum. The training set contains numbers with up to 8 digits, and the test set contains numbers with 9 or 10 digits. Numbers are padded  to reach a length of 12 so that it's easy to align the digits that need to be added. We found that without padding, the task became much harder.
    {\bf Types of compositionality:} models need to learn that there is a primitive operation ``adding two digits (with carry)'' that is repeatedly applied at each position. Models that learn position-specific shortcuts will not generalize to longer input lengths (as they would have learned no rules to produce the most significant digits, which would have never been seen during training). This mostly corresponds to {\em productivity} type of compositional generalization.
    
    \item 
    {\bf AdditionNegatives} ({\em AddNeg}): The same as the previous one, but 25\% of the numbers are negative (preceded with the ``\texttt{-}'' token). {\bf Types of compositionality:} the type of compositionality requires by this task is similar to that of the previous task, except that the general rules that need to be learned (independent of position) are more complex due to negative numbers. So, the model needs to learn three basic primitive operations that are the same irrespective of the position of the digits: ``add two digits with carry'', ``subtract first from second with carry'', and ``subtract second from first with carry'', and learn when to apply each. This also mostly corresponds to {\em productivity} type of compositional generalization. 
    
    \item 
    {\bf Reversing} ({\em Reverse}): Where the output is expected to be the input sequence in reverse order. 
    Training contains sequences of up to 16 digits, and the test set contains lengths between 17 to 24. {\bf Types of compositionality:} the difficult part of this task is to learn to reverse position embeddings in a way that generalizes to longer inputs than seen during training, in order to attend and produce the right output sequences. This mostly corresponds to {\em productivity} type of compositional generalization, as the model needs to learn to reverse position embeddings for longer sequences than seen during training.
    
    \item 
    {\bf Duplication} ({\em Dup}): The input is a sequence of digits and the output should be the same sequence, repeated twice. 
    Training contains sequences up to 16 digits, and test from 17 to 24. {\bf Types of compositionality:} Learning to repeat the input several times is not a particularly hard task for a Transformer, but we noticed that the difficult part was learning when to stop producing output (exactly after repeating the input twice in this case). This problem was also noted in the work of \cite{csordas2021devil}, and mostly corresponds to {\em productivity} type of compositional generalization.
    
    \item 
    {\bf Cartesian} ({\em Cart}): The input contains two sequences of symbols, and the output should be their Cartesian product. 
    Training contains sequences of up to 6 symbols (7 or 8 for testing). {\bf Types of compositionality:} this is a very challenging task that requires very demanding {\em productivity}, as the model needs to learn to learn to compose the basic operation of pairing elements from both sets via two nested loops: iterating over each of the two input sets. 
    
    \item 
    {\bf Intersection} ({\em Inters}): Given two sequences of symbols, 
    the output should be whether they have a non-empty intersection. Training contains sets with size 1 to 16, and testing 17 to 24. {\bf Types of compositionality:} the main challenge in this dataset is to learn short-cut rules such as ``if the first set contains \texttt{a4} and the second set also contains \texttt{a4} then the output should be \texttt{true}''. However, the model needs to learn to ignore these token specific rules, and learn the general rule of finding two identical tokens regardless of which specific token they are, which could be seen as a form of {\em systematicity}. Moreover, this needs to be learned in a way that generalizes to longer inputs ({\em productivity}).
    
    \item 
    {\bf SCAN-length} ({\em SCAN-l}): The {\em length split} of the SCAN dataset~\cite{lake2018generalization}. The SCAN dataset asks the model to learn to interpret and execute natural language instructions with a limited vocabulary. For example, if the input is ``walk twice", the output should be ``I\_WALK I\_WALK``. There are a set of primitive actions (walk, jump, etc.), and a set of modifiers (twice, thrice, left, etc.) and composition operators (e.g., and), and the model needs to learn how to compose and execute all of those instructions to generate the output sequence. In this specific {\em length split}, the training and test sets are split by length (the test set contains the longest sequences and the training set the shortest ones). {\bf Types of compositionality:} Overall, SCAN requires significant {\em systematicity} to be solved, and this split in particular focuses on {\em productivity}.
    
    \item 
    {\bf SCAN-add-jump} ({\em SCAN-aj}): The {\em add primitive jump split} of the SCAN dataset~\cite{lake2018generalization}. In this split, the ``jump'' instruction is only seen during training in isolation (i.e., there is a training example ``jump'' $\rightarrow$ ``I\_JUMP''), but the test set contains this instruction heavily, and in combination with other constructs. {\bf Types of compositionality:} this split in particular focuses more on {\em systematicity}.
    
    \item 
    {\bf PCFG-productivity} ({\em PCFG-p}): The productivity split of the PCFG dataset~\cite{hupkes2020compositionality}. The PCFG dataset is a synthetic dataset where each example contains a set of operations that need to be done to one or more input strings, and the model needs to learn to apply these operations and produce the final output. Operations include reversing, duplicating, getting the first element, etc. {\bf Types of compositionality:} this split in particular focuses on {\em productivity}, as test examples contain longer sequences of instructions than those seen during training.
    
    \item 
    {\bf PCFG-sytematicity} ({\em PCFG-s}: The systematicity split of the PCFG dataset~\cite{hupkes2020compositionality}. {\bf Types of compositionality:} this split focuses on {\em systematicity}, by testing the model recombining operations in ways never seen during training.
    
    \item 
    {\bf COGS}: The generalization split of the COGS semantic parsing dataset~\cite{kim2020cogs}. This is a semantic parsing dataset, where the input is a sentence in natural language, and the output should be a logical representation of the sentence. {\bf Types of compositionality:} The generalization split contains combinations not seen during training, while most of these focus on {\em systematicity} (e.g., constructions that had only been seen as subjects, now they are seen as objects), some part of the test set focuses on {\em productivity} (having deeper nesting of propositional phrases, for example). This, productivity type of generalization, is where our sequence tagging approach significantly outperforms previous approaches.
    
    \item 
    {\bf CFQ-mcd1} ({\em CFQ}): The MCD1 split of the CFQ dataset~\cite{keysers2019measuring}. This dataset asks a model to learn how to translate delexicalized natural language queries into SPARQL. {\bf Types of compositionality:} the MCD1 split of this dataset focuses specifically on {\em systematicity}, but more concretely, there are two additional ways in which this dataset is hard compositionally. First, solving this dataset requires solving Cartesian products (which is the reason for which we added the separate Cartesian product task), since some question contains constructions like: ``Who directed, played and produced movies M1, M2 and M3'', which get translated into 9 SPARQL clauses (the Cartesian product). Second, SPARQL clauses are supposed to be produced in alphabetical order, hence the model needs to learn how to sort.
\end{itemize}

Finally, table \ref{tbl:datasets} shows the size of the training and test sets for each dataset, as well as the size of their vocabularies. For the vocabulary, we used the union of the input and output vocabularies as a unified vocabulary.
We also show the number of training epochs we performed in each dataset (this was chosen as the number after which performance stabilized with some initial models; it was not tuned afterwards during the systematic evaluation presented below).

\end{document}